\documentclass{article}

\PassOptionsToPackage{numbers,compress}{natbib}
\usepackage[dblblindworkshop,final]{neurips_2026}

\usepackage[utf8]{inputenc}
\usepackage[T1]{fontenc}
\usepackage{hyperref}
\usepackage{url}
\usepackage{booktabs}
\usepackage{amsfonts}
\usepackage{nicefrac}
\usepackage{microtype}
\usepackage{xcolor}
\usepackage{graphicx}
\usepackage{subcaption}
\usepackage{amsmath}
\usepackage{pifont}

\workshoptitle{DiffuLM @ NeurIPS 2026}

\title{Label Collapse in Diffusion Language Model In-Context Learning}

\author{%
  Ziye Guo\textsuperscript{1} \\
  \textsuperscript{1}University of Connecticut \\
  \texttt{qxl26001@uconn.edu}
  \And
  Ye Yuan\textsuperscript{2} \\
  \textsuperscript{2}McGill University \\
  \texttt{ye.yuan3@mail.mcgill.ca}
}

\begin{document}

\maketitle

\begin{abstract}
In-context learning (ICL) applies large language models to new tasks by prepending a few labeled demonstrations to a query.
Prior work has studied how demonstration ordering affects ICL, but the interaction between label arrangement patterns and model biases has not been systematically examined.
We define \emph{label collapse} as the failure mode in which the model generates query-independent labels, and show that it is architecturally universal at the tested $8$--$9$B scale: five models spanning autoregressive (Llama-3, Qwen3.5) and discrete diffusion (LLaDA, iLLaDA, LLaDA-Instruct) families all exhibit label collapse \emph{even under} the class-balanced alternating arrangement, with some label word pairs collapsing in virtually every tested condition and others flipping their collapse direction with arrangement.
Critically, accuracy alone cannot detect this collapse: models can achieve near-chance accuracy while generating query-independent labels, a score indistinguishable from balanced performance, so the failure can persist undetected in deployment.
We measure \emph{separation}, the log-odds of the two label words at the query position (the quantity that label-bias calibration operates on), as a complementary diagnostic that exposes collapse hidden beneath surface accuracy, together with a measurement protocol that guards label extraction against tokenizer and formatting artifacts.
For autoregressive models, collapse persists where the majority-label-bias account does not apply, and arrangement patterns can reverse the collapse direction; for diffusion models, we present a systematic evaluation of ICL failure modes, complementing contemporaneous studies of dLLM context comprehension and demonstration-order sensitivity, and identify open questions about how iterative denoising interacts with label biases.
\end{abstract}

\section{Introduction}
\label{sec:intro}
In-context learning (ICL), the ability to perform new tasks from demonstrations in the prompt without weight updates, is a foundational capability of modern language models~\citep{gpt3,dong2022survey}.
A natural approach to class-balanced ICL prompts is an alternating label order.
Prior work has studied how demonstration ordering affects ICL performance~\citep{lu2022fantastically,zhao2021calibrate}, but the interaction between label arrangement patterns and model biases has not been systematically examined.
The gap is particularly acute for diffusion language models (dLLMs), whose ICL capabilities have only recently been demonstrated~\citep{llada}; contemporaneous work has begun to characterize their context comprehension and demonstration-order sensitivity~\citep{piskorz2026masks}, but query-independent label failure has not been isolated in dLLMs, nor is it known whether AR label-bias findings transfer to iterative denoising.

We define \emph{label collapse} as a failure mode in which the model generates query-independent labels: regardless of the query input, the model outputs the same label token.
We study it along four axes: (1)~five models (three dLLMs and two autoregressive (AR) baselines; Section~\ref{sec:setup}), (2)~20 label word pairs spanning sentiment, safety, geometric, and abstract domains, (3)~four arrangements of the same demonstrations, and (4)~demonstration counts $K \in \{4,\ldots,30\}$.

We show that label collapse is \emph{architecturally universal but graded}: every model collapses on some label pairs, with the collapse direction largely traceable to a token-level prior rather than determined by architecture.
Critically, accuracy alone cannot distinguish this collapse from balanced competence: a collapsed model that always outputs the same label word scores $\sim$50\% on a balanced binary task, and the failure persists undetected because standard protocols never probe whether predictions are query-dependent.

To expose it, we measure \emph{separation}: the log-odds of the positive and negative label words at the query position, computed from a single forward pass.
This quantity underlies label-bias calibration~\citep{zhao2021calibrate,holtzman2021surface,zhou2023batch}, which subtracts reference points from it to correct predictions; we instead read its distribution over queries as a diagnostic of collapse.
Unlike accuracy, separation directly measures whether the model's internal representations distinguish between query inputs.
We further provide a measurement protocol for reliable label extraction:
tokenizers may split a label word into multiple tokens, and output formats may shift the answer off the parsed line, so we audit every condition for both failure modes and flag the conditions they confound.

In summary, our contributions are:
(1)~We identify label collapse and demonstrate its architectural universality across five AR and dLLM models, 20 label word pairs, and four arrangements, with severity ranging from universal collapse to model-dependent graded failure.
(2)~Accuracy alone cannot detect this collapse: collapsed models achieve near-chance accuracy that is indistinguishable from balanced performance, so detection requires a model-internal diagnostic.
(3)~We adopt separation (the label log-odds at the query position, the quantity underlying label-bias calibration) as a model-agnostic single-pass diagnostic that exposes collapse beneath surface accuracy, with a measurement protocol that guards label extraction against tokenizer and formatting artifacts; the bias structure it exposes also predicts which collapsed conditions calibration can and cannot repair.
(4)~Arrangement can reverse collapse direction for weak-bias pairs but cannot avert strong-bias collapse; how iterative denoising in dLLMs interacts with label biases remains open.
We review related work on label biases, demonstration order, and diffusion language models in Appendix~\ref{sec:related}.

\section{Experimental Setup}
\label{sec:setup}

\paragraph{Models.}
We evaluate five models: three diffusion and two autoregressive language models.
\textbf{(i)}~For diffusion language models, we use (1)~LLaDA-8B-Base~\citep{llada}, (2)~iLLaDA-8B-Base~\citep{illada}, independently pre-trained from scratch on 12T tokens and \emph{not} an SFT (supervised fine-tuning) derivative of LLaDA, and (3)~LLaDA-8B-Instruct~\citep{llada}, chat-tuned from LLaDA-Base (the only same-lineage SFT pair).
\textbf{(ii)}~For autoregressive language models, we use (4)~Llama-3-8B-Instruct~\citep{llama3} and (5)~Qwen3.5-9B-Base~\citep{qwen35}.
All dLLMs use \texttt{gen\_length=3, steps=8, block\_length=3, temperature=0}; AR models use greedy decoding with \texttt{max\_new\_tokens=10}.

\paragraph{Tasks and Label Word Pairs.}
We use two classification datasets: SST-2~\citep{socher2013sst2} for sentiment classification and XSTest~\citep{rottger2023xstest} for safety classification.
We define 20 label word pairs across four semantic domains. \textbf{(i)}~Sentiment: (1)~positive/negative, \textbf{(ii)}~Safety: (2)~clean/toxic and (3)~safe/unsafe, \textbf{(iii)}~Geometric: (4)~circle/square, \textbf{(iv)}~Adjective and Abstract Domains: (5)~true/false, (6)~yes/no, (7)~cat/dog, (8)~hot/cold, (9)~correct/wrong, (10)~right/left, (11)~easy/hard, (12)~fast/slow, (13)~strong/weak, (14)~large/small, (15)~light/dark, (16)~ancient/modern, (17)~kind/cruel, (18)~quiet/loud, (19)~rich/poor, and (20)~wet/dry.
Pairs (1)--(10) are \emph{task pairs} and (11)--(20) \emph{antonym pairs}.
All prompts use a fixed template in which each demonstration is a sentence followed by its label word after ``Answer:'', and the prompt ends with ``Answer:'' after the query, with no trailing space.
A complete prompt example is given in Appendix~\ref{app:details}.

\paragraph{Evaluation Metrics.}
We evaluate two modes:
(1)~\textbf{Generation (gen)}: the model generates the label token; we measure it by \texttt{pos\_rate} (fraction of outputs matching the positive label) and by accuracy (an output is correct only if it matches the query's gold label word; invalid outputs, matching neither label, are incorrect);
(2)~\textbf{Single-pass (sp)}: from one forward pass we read the logits of the two label words $w_{+}$ and $w_{-}$ at the masked (answer) position; their difference is the \emph{separation},
\begin{equation}
\label{eq:sep}
\mathrm{sep}(x) \;=\; \mathrm{logit}(w_{+} \mid x) - \mathrm{logit}(w_{-} \mid x) \;=\; \log \frac{P(w_{+} \mid x)}{P(w_{-} \mid x)},
\end{equation}
abbreviated \emph{sep} in formulas ($x$: the query together with its demonstrations); the second equality is exact because the softmax denominators cancel.
The single-pass \texttt{pos\_rate} is the fraction of queries with $\mathrm{sep}(x) > 0$, and single-pass accuracy is the fraction whose argmax label matches the query's gold label word.
Separation is thus the label log-odds, the same quantity that label-bias calibration methods correct against reference points (contextual calibration~\citep{zhao2021calibrate}, PMI scoring~\citep{holtzman2021surface}, and batch calibration~\citep{zhou2023batch}); we use it, without correction, as a diagnostic (Section~\ref{sec:mechanism}).
We use four severity terms with explicit membership conditions, all stated in terms of \texttt{pos\_rate} and accuracy on the class-balanced query sets.
(1)~\emph{Hard collapse}: \texttt{pos\_rate} $\leq 0.05$ or $\geq 0.95$ \emph{and} accuracy at chance ($\leq 0.55$), observed in every tested condition of a model--pair combination; the model emits one label word for virtually every query.
(2)~\emph{Collapse}: \texttt{pos\_rate} $\leq 0.20$ or $\geq 0.80$ with accuracy near chance ($\leq 0.60$).
(3)~\emph{Drift}: \texttt{pos\_rate} at or beyond $\pm 0.15$ of balance while accuracy remains well above chance ($\geq 0.70$); drift is \emph{functional} when accuracy lies within $0.05$ of the ceiling $0.5 + \min(p, 1-p)$ imposed by its own skew ($p$~denotes \texttt{pos\_rate}).
(4)~\emph{Healthy}: \texttt{pos\_rate} strictly within $\pm 0.15$ of balance and accuracy $\geq 0.85$.

The thresholds are anchored to the observed distribution rather than tuned (hard collapse's $0.05$ boundary sits just above the largest \texttt{pos\_rate} of any hard-collapse condition, $0.03$) and are stable under perturbation: shifting a single boundary by $\pm 0.05$ reclassifies at most $7.0\%$ of conditions, leaving the median grade unchanged.

\paragraph{Demonstration Configurations.}
We use four arrangements of the same demonstration pool: \emph{alternating} (PNPN), \emph{grouped} (PPNN), \emph{blocked} (PPPNNN), and \emph{shuffled} (a per-$K$ random permutation), where P and N denote the positive and negative labels; all four order the same label multiset, $\lceil K/2\rceil$ positives and $\lfloor K/2\rfloor$ negatives at each $K$ (constructions in Appendix~\ref{app:details}).
Under alternating we sweep the number of demonstrations $K$ from 0 to 30 for the two dLLM base models and Llama-3 (all integers), and from 0 to 13 plus $K \in \{15, 16, 20, 24, 30\}$ for Qwen3.5-Base and LLaDA-Instruct; counts with $K < 4$, reported in the exact grids of Appendix~\ref{app:details}, enter only the zero-shot baselines described below and the neutral-answer analysis of Appendix~\ref{app:neutral}; the other arrangements are spot-checked at $K \in \{4, 8, 12, 16, 30\}$, the common even-$K$ grid on which every arrangement is exactly class-balanced (exact grids in Appendix~\ref{app:details}), and a continuous $K{=}5$--$14$ grouped sweep covers the two dLLM base models on the four representative pairs.
The grids differ across models for compute reasons; all cross-model snapshots use $K{=}8$, which was evaluated for every model, with a single flagged exception.
As an unconditional baseline, $K{=}0$ (template + query only) is evaluated for the ten task pairs across all five models in both modes, and for the ten antonym pairs in the single-pass mode.
Each condition uses 200 query sentences.
\section{Label Collapse Phenomenon}
\label{sec:phenomenon}

\begin{figure}[t]
  \centering
  \includegraphics[width=\textwidth]{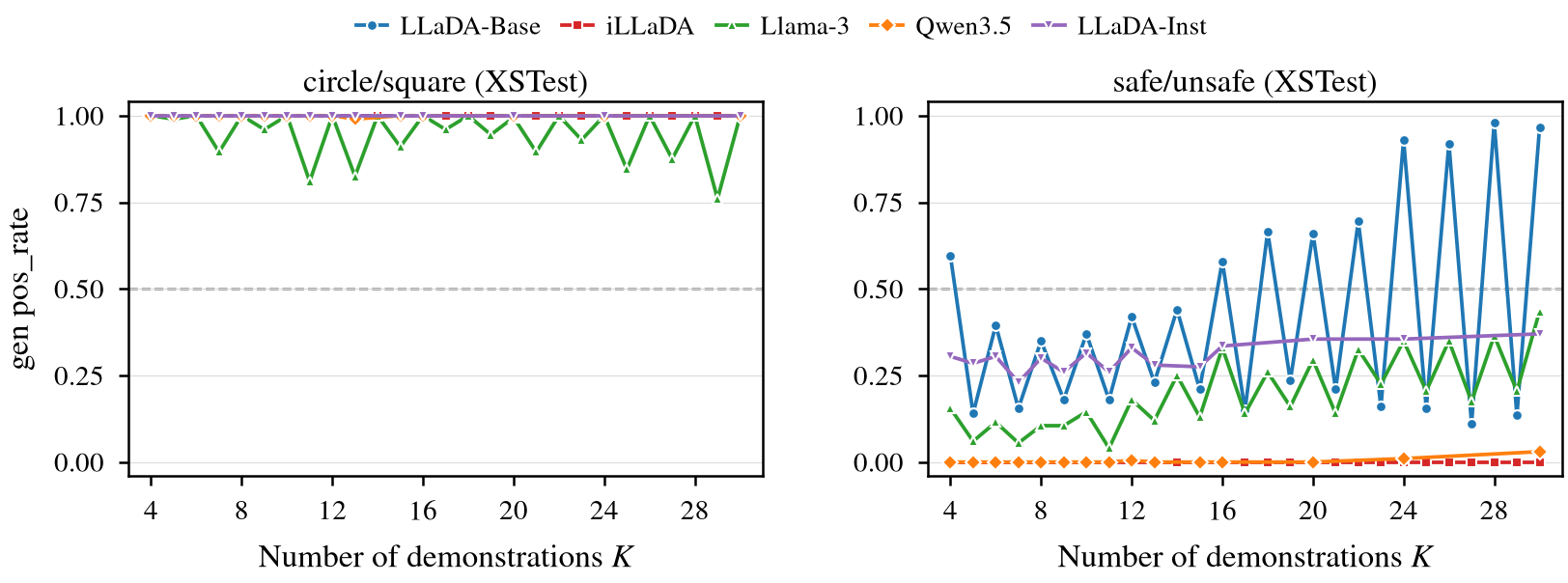}
  \caption{K-sweep generation \texttt{pos\_rate} for the two anchor pairs (circle/square, safe/unsafe; alternating arrangement). Gray dashed line = 0.5 (balanced). circle/square collapses to 1.0 for all models at every $K$; on safe/unsafe, iLLaDA collapses to 0.0 while Llama-3 gradually recovers toward 0.5 as $K$ grows and LLaDA-Base reverses direction at high $K$. The remaining two representative pairs (positive/negative, clean/toxic) are shown in Appendix~\ref{app:details}.}
  \label{fig:ksweep-orig4}
\end{figure}

\begin{table}[t]
\centering
\small
\caption{Generation \texttt{pos\_rate} for all 20 word pairs at $K{=}8$ under alternating (200 queries per
cell), sorted into severity tiers. Tier assignment uses all arrangements and $K$
values tested (\emph{not} only this snapshot): ``universal'' collapses under
every arrangement/$K$; ``healthy'' stays functional for all five models.
\ding{55}~=~collapse ($|\text{\texttt{pos\_rate}}-0.5|\geq0.3$); $\triangle$~=~drift
($0.15\leq|\text{\texttt{pos\_rate}}-0.5|<0.3$); $^{\dagger}$~=~more than 30\% of
outputs leave the label vocabulary or the answer format; such cells are excluded from
collapse counts (Appendix~\ref{app:bpe}). The ancient/modern row reflects the alternating arrangement only: under grouped and shuffled its cells reverse toward ``modern'', and under blocked the two dLLM base models return to balance (Section~\ref{sec:arrangement}).}
\label{tab:tier-map}
\begin{tabular}{lccccc}
\toprule
Word pair & LLaDA & iLLaDA & LLaDA-Inst. & Llama-3 & Qwen3.5 \\
\midrule
\multicolumn{6}{l}{\emph{Universal-collapse}} \\
circle/square & 1.00\,\ding{55} & 1.00\,\ding{55} & 1.00\,\ding{55} & 1.00\,\ding{55} & 1.00\,\ding{55} \\
\midrule
\multicolumn{6}{l}{\emph{Healthy (all five models functional)}} \\
strong/weak & 0.49 & 0.48 & 0.47 & 0.59 & 0.48 \\
\midrule
\multicolumn{6}{l}{\emph{Model-dependent}} \\
positive/negative & 0.89\,\ding{55} & 0.50 & 0.90\,\ding{55} & 0.91\,\ding{55} & 0.88\,\ding{55} \\
clean/toxic & 1.00\,\ding{55} & 0.43 & 1.00\,\ding{55} & 0.61 & 0.46 \\
safe/unsafe & 0.35\,$\triangle$ & 0.00\,\ding{55} & 0.30\,$\triangle$ & 0.10\,\ding{55} & 0.00\,\ding{55} \\
true/false & 0.41 & 0.00\,\ding{55} & 0.96\,\ding{55} & 0.88\,\ding{55} & 0.58 \\
yes/no & 0.46 & 0.48 & 0.57 & 1.00\,\ding{55} & 1.00\,\ding{55} \\
cat/dog & 0.58 & 0.53 & 0.82\,\ding{55} & 1.00\,\ding{55} & 0.59 \\
hot/cold & 0.99\,\ding{55} & 0.48 & 1.00\,\ding{55} & 0.98\,\ding{55} & 0.99\,\ding{55} \\
correct/wrong & 0.49 & 0.54 & 0.65\,$\triangle$ & 0.78\,$\triangle$ & 0.51 \\
right/left & 1.00\,\ding{55} & 0.20$^{\dagger}$\,\ding{55} & 1.00\,\ding{55} & 0.01$^{\dagger}$\,\ding{55} & 0.48 \\
easy/hard & 0.97\,\ding{55} & 0.96\,\ding{55} & 0.47 & 0.60 & 0.52 \\
fast/slow & 1.00\,\ding{55} & 0.49 & 1.00\,\ding{55} & 0.95\,\ding{55} & 0.53 \\
kind/cruel & 1.00\,\ding{55} & 0.48 & 1.00\,\ding{55} & 0.99\,\ding{55} & 1.00\,\ding{55} \\
large/small & 1.00\,\ding{55} & 0.62 & 0.51 & 0.97\,\ding{55} & 0.48 \\
light/dark & 0.99\,\ding{55} & 0.49 & 1.00\,\ding{55} & 0.54 & 0.51 \\
quiet/loud & 1.00\,\ding{55} & 0.55 & 1.00\,\ding{55} & 1.00\,\ding{55} & 0.85\,\ding{55} \\
rich/poor & 1.00\,\ding{55} & 0.48 & 1.00\,\ding{55} & 0.99\,\ding{55} & 0.50 \\
wet/dry & 0.99\,\ding{55} & 0.48 & 1.00\,\ding{55} & 0.31$^{\dagger}$\,$\triangle$ & 0.66\,$\triangle$ \\
ancient/modern & 0.93\,\ding{55} & 0.65\,$\triangle$ & 1.00\,\ding{55} & 0.98\,\ding{55} & 0.55 \\
\bottomrule
\end{tabular}
\end{table}

\paragraph{A three-tier severity structure.}

\textbf{Which words occupy the label space, not the architecture, governs the severity of collapse.}
Figure~\ref{fig:ksweep-orig4} shows the K-sweep generation \texttt{pos\_rate} for the two anchor pairs, and the full 20-pair grid (Table~\ref{tab:tier-map}) resolves into three tiers.
(i)~\emph{Universal collapse}: circle/square stays above $0.80$ (toward ``circle'') in 170/172 generation conditions with $K \geq 4$, and the direction is fixed before any demonstration: at $K{=}0$ the single-pass prior of all five models already leans toward ``circle'' (mean positive probability $0.70$--$0.83$), consistent with their unigram logits (Section~\ref{sec:mechanism}).
(ii)~\emph{Healthy}: strong/weak is the only pair functional for all five models, and correct/wrong is intermediate: three models stay healthy while two drift toward ``correct''.
(iii)~\emph{Model-dependent}: safe/unsafe spans the full spectrum, from hard collapse (iLLaDA and Qwen3.5: \texttt{pos\_rate} $\leq 0.03$ at every tested $K$ and arrangement) through steady recovery (Llama-3) to a parity-modulated reversal (LLaDA-Base, whose even-$K$ flips toward ``safe'' trace the extra positive carried by odd-length alternating blocks; Appendix~\ref{app:parity}).

\paragraph{Lineage divergence on the antonym pairs.}

\textbf{The two dLLM lineages have nearly disjoint dangerous-pair sets.}
The ten antonym pairs reuse the SST-2 demonstration pool of the positive/negative pair with only the label words exchanged, so label-word identity is the sole manipulated variable.
At $K{=}12$ blocked, the two dLLM base models sit in opposite regimes (the AR baselines were not evaluated at this condition): (i)~LLaDA-Base collapses on 6 of the 10 pairs, all toward the positive word (fast, kind, light, quiet, rich, wet); (ii)~iLLaDA collapses on none (\texttt{pos\_rate} $0.46$--$0.51$ across all ten); (iii)~LLaDA-Instruct tracks LLaDA-Base and adds ancient/modern, reaching 7 of 10.
The contrast persists under alternating at every $K$ (LLaDA-Base: 5 of 10 pairs collapsed at $K{=}4$, 9 at $K{=}12$; iLLaDA: at most 3 of 10 at any $K$).
LLaDA-Base collapses on the antonym pairs while retaining partial function on safe/unsafe, and iLLaDA shows the reverse (attribution in Section~\ref{sec:sft}).

\paragraph{A surprising inversion on the task's own label pair.}
\textbf{On positive/negative, iLLaDA is the only healthy model, and the repair is not a weaker prior.}
Across its 51 arrangement--$K$ conditions with $K \geq 4$, iLLaDA never collapses (\texttt{pos\_rate} $0.38$--$0.56$, accuracy $0.87$--$0.96$), while the other four models lean toward ``positive'' under alternating at every tested $K \geq 4$ ($0.69$--$1.00$ pooled, accuracy $0.48$--$0.52$; Qwen3.5: $0.50$--$0.78$); Llama-3 alone drifts back toward ``negative'' under grouped at large $K$.
The $K{=}0$ priors do not explain the split: iLLaDA's prior ($0.70$) is as positive-lean as LLaDA-Base's ($0.77$), yet demonstrations suffice to override it in iLLaDA but fail to do so in LLaDA-Base, although large AR models are known to override such priors given flipped-label demonstrations~\citep{wei2023larger}.
The repair lies in the capacity of demonstrations to override the prior, a capacity acquired from a different pre-training corpus and not conferred by chat SFT (LLaDA-Instruct: $K{=}0$ prior $0.56$, yet collapsed by $K{=}4$; Section~\ref{sec:sft}).

\paragraph{Hard collapse is indifferent to arrangement.}
\textbf{When the token-level bias is strong enough, no arrangement rescues the model.}
Circle/square stays pinned to ``circle'', and safe/unsafe stays pinned to ``unsafe'' for iLLaDA (\texttt{pos\_rate} $0.000$ in all 51 conditions) and Qwen3.5 ($\leq 0.03$ in all 24; both with $K \geq 4$), across all four arrangements and every tested $K$.
This bias is insensitive to demonstration ordering but not to the demonstrations' presence: under the logit lens~\citep{nostalgebraist2020logit} it is read out at the final output projection layer (Section~\ref{sec:mechanism}), and iLLaDA's $K{=}0$ prior leans toward ``safe'' ($0.61$), so adding demonstrations flips it toward ``unsafe''.
The remaining models on safe/unsafe are arrangement- and $K$-dependent (Section~\ref{sec:arrangement}): LLaDA-Base collapses under grouped (13/13 conditions) yet merely drifts under shuffled ($0.23$--$0.41$), and LLaDA-Instruct stays functional under alternating and blocked at every tested $K \geq 4$ but slides into collapse under grouped at large $K$ ($0.20 \rightarrow 0.05$ from $K{=}12$ to $K{=}30$).

\paragraph{Effect of demonstration count.}
\textbf{More demonstrations are not uniformly beneficial.}
On safe/unsafe under alternating, they can lock a bias in (iLLaDA and Qwen3.5 keep separation below $-3$ at every $K$), slowly repair it (Llama-3 climbs back toward zero), or flip its sign (LLaDA-Base reverses at large $K$), in contrast to the monotonic gains reported for many-shot ICL~\citep{agarwal2024manyshot}.
Several of these behaviors are parity-structured (odd vs.\ even $K$ changes both the terminal label and the class balance of an alternating prompt), always in the direction opposed to recency and majority bias, and never under grouped (Appendix~\ref{app:parity}).

\paragraph{AR models also collapse, with signatures not shared by the dLLMs.}
\textbf{Both AR models collapse broadly: at $K{=}8$ alternating, Llama-3 collapses on 13 of its 18 parseable pairs and Qwen3.5 on 7 of 20.}
Their signatures differ from those of the dLLMs in three ways.
(i)~\emph{Escape from the label vocabulary}: on right/left, Qwen3.5 answers largely outside the pair at small $K$ (out-of-pair rate $45$--$74\%$) and returns to it at large $K$, remaining balanced within it ($0.38$--$0.49$), a drift \emph{into} the label space rather than toward a label word.
(ii)~\emph{Parity oscillation that reaches behavior}: on true/false, Llama-3's separation oscillates by parity (even $K$: $+3.0$ to $+4.0$; odd: near zero) and its generation follows (even $K \geq 10$: $0.79$--$0.91$ toward ``true''; odd: $0.28$--$0.44$); under grouped it collapses toward ``true'' up to $K{=}16$ and drifts back at $K{=}30$.
(iii)~\emph{Parsing artifacts}: the label word may split into multiple tokens (``toxic'' as \texttt{to}+\texttt{xic}, among others), so first-token matching misreads correct outputs as unparseable; we parse outputs at the text level throughout, which restores these conditions (Appendix~\ref{app:bpe}).
\section{Mechanism: Separation as Diagnostic}
\label{sec:mechanism}
This section localizes the bias inside the network, traces it to token priors, and examines when the single-pass channel faithfully predicts generation.

\paragraph{Logit lens localization.}
\textbf{In all cases the bias is set at or before the final projection layer (\texttt{lm\_head}); the intermediate contextual layers do not repair it.}
Applying the logit lens to all 33 layers (L0--L32; L0 is the embedding layer) of LLaDA-Base, iLLaDA, and Llama-3 (all ten task pairs, alternating, $K{=}8$), we find two patterns.
For Llama-3 and iLLaDA, label separation stays near zero through the intermediate layers and emerges only in the final layers (peak at L29--L32).
For LLaDA-Base the picture is heterogeneous: several pairs (clean/toxic, safe/unsafe, true/false, circle/square) already show large separation at layer 0, consistent with a token-embedding-level bias, while others (positive/negative, hot/cold, cat/dog) emerge late (peak at L26--L29).
This heterogeneity motivates measuring the bias with no context at all.

\paragraph{Zero-shot ($K{=}0$) baseline and unigram logits.}
\textbf{Roughly three quarters of collapse directions trace back to a token-level prior readable at the output head before any demonstration is seen.}
We measure the label prior with no demonstrations in two independent ways, both reading the label tokens themselves rather than the demonstrations: we call this a \emph{token-level prior}.
The zero-shot single-pass evaluation uses the query followed by the answer slot (\texttt{<query> $\backslash$n Answer:}), so the two label logits are read at the position where a demonstration-conditioned model would write its answer.
Across all 20 word pairs and five models (100 conditions), the sign of the zero-shot separation agrees with the high-$K$ collapse direction (median \texttt{pos\_rate} over the pair's $K \geq 4$ conditions) in 33/45 collapsed model--pair combinations (73\%).
A complementary measurement drops the query as well: empty-context unigram logits at the output head (prompt reduced to the bare answer slot; 18 of 100 pairs skipped for multi-token labels, Appendix~\ref{app:bpe}) agree in 28/37 collapsed cases (76\%).
The two measurements cross-validate each other (Appendix~\ref{app:calib}), whereas a pure word-frequency account (predicting that the more frequent label word wins~\citep{holtzman2021surface}) matches the collapse direction in only 7/14 cases where the two label words differ by at least a factor of $2.5$ in corpus frequency: the label prior is model-internal, not a corpus-frequency effect.
Disagreements survive where the prior is weak or opposed (Llama-3 wet/dry leans ``wet'' at $K{=}0$, $+2.07$, yet collapses to ``dry''); the remainder are demonstration-induced flips.
Three fresh word pairs, constructed after these analyses and matched in token length (equal token counts for the two label words under all five tokenizers), replicate the pattern: collapse survives equal lengths, and the prior-direction exceptions again trace to a single lineage (Appendix~\ref{app:newpairs}).
Standard calibrations subtract a reference point from the log-odds (Appendix~\ref{app:calib}): on the single-pass channel at $K{=}8$ (86 conditions, 45 of them collapsed), contextual calibration~\citep{zhao2021calibrate} repairs 16/45 and query-conditional PMI~\citep{holtzman2021surface} repairs 4/45, whereas batch calibration~\citep{zhou2023batch}, which subtracts the test-batch mean, returns all 45 collapsed conditions to balance and restores accuracy in 36 of them.
The 45/45 result is the formal signature of query-independent bias (one shared offset per condition suffices to restore balance), while the nine residual conditions carry no query information after centering, and eight of them are reached by no reference-point correction at all (Appendix~\ref{app:calib}); batch calibration is transductive, so it locates the bias rather than fixes deployment.
Among hard-tier conditions, repair is predicted by the prior share $|\log W|/|\mathrm{sep}|$ ($W{=}W_{\mathrm{pos}}/W_{\mathrm{neg}}$ from content-free prompts): below $0.55$ never ($0/18$), at or above $0.8$ repaired in four of five (the single exception sits at $0.80$), and the boundary runs between model--pair combinations rather than word pairs (Appendix~\ref{app:calib}).
Injecting the correction into dLLM decoding repairs isolated conditions (LLaDA-Base positive/negative: $0.89 \rightarrow 0.51$) but leaves the pinned word pairs (circle/square, true/false) unmoved, with occasional overshoot (Appendix~\ref{app:calib}).

\paragraph{Separation as a behavioral predictor.}
\textbf{A single forward pass identifies behavioral collapse with high precision, and the predictor's failure modes are themselves informative.}
Across the 1{,}502 paired conditions, the criterion $\lvert\mathrm{sep}\rvert > 3$ identifies behavioral collapse ($\lvert\texttt{pos\_rate} - 0.5\rvert \geq 0.3$) with 92.5\% precision and 84.7\% recall.
Varying the threshold between 2 and 5 and the behavioral margin between 0.2 and 0.4 keeps precision within 77--99\% and recall within 52--95\%; we fix $3$ as a conservative default that favors precision.
False positives concentrate on LLaDA-Instruct safe/unsafe (17 of 64), where the single pass anticipates a collapse that the iterative process never commits to, and on partially recovered Llama-3 pairs (22); false negatives concentrate where generation has already flipped while the single-pass bias lags: the true/false decoupling (15), parity-lagged safe/unsafe flips (21, 13 at odd $K$), and the even-$K$ arm of iLLaDA clean/toxic (9).

\begin{figure}[t]
\centering
\includegraphics[width=\textwidth]{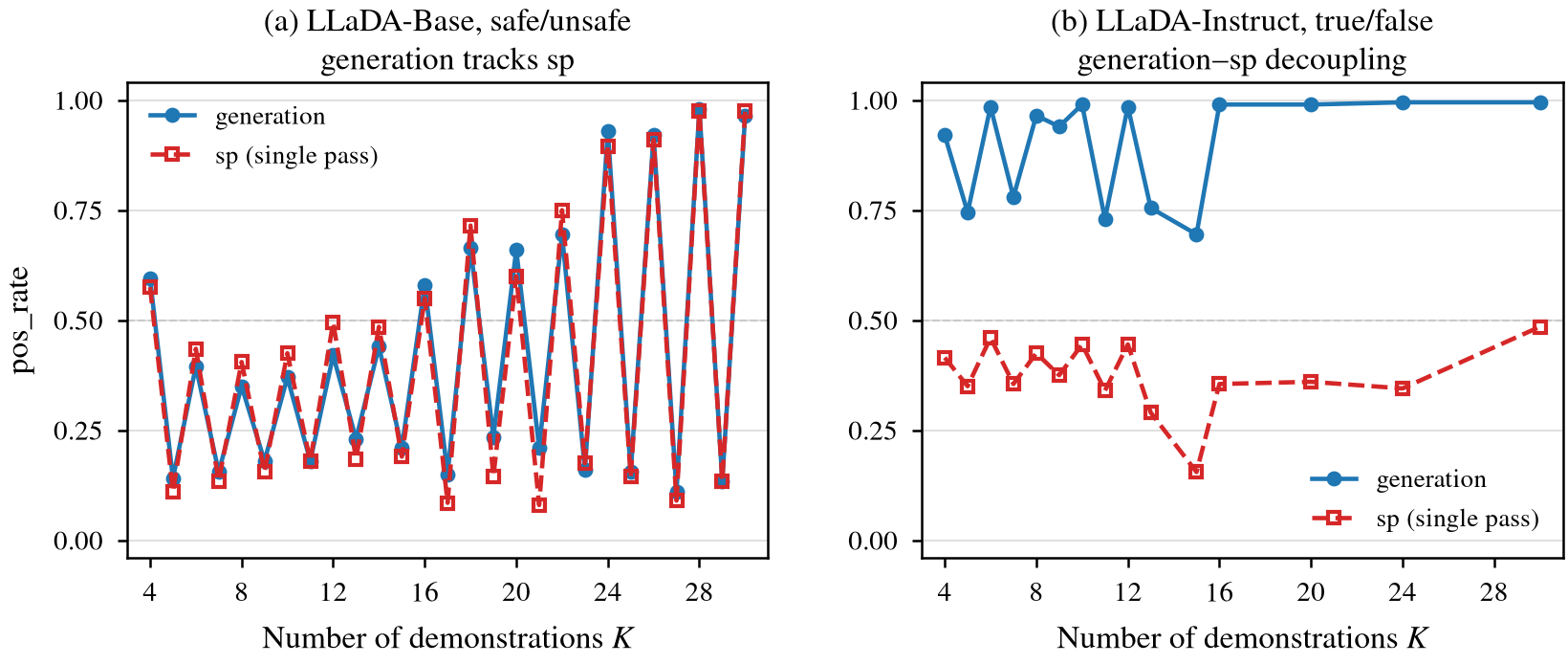}
\caption{Generation vs.\ single-pass (sp) \texttt{pos\_rate} on the same 200 queries per condition (alternating arrangement).
(a)~LLaDA-Base, safe/unsafe, $K{=}4$--$30$: the two channels track each other point by point (max difference $0.13$).
(b)~LLaDA-Instruct, true/false, $K{=}4$--$30$: generation collapses to ``true'' ($0.69$--$0.99$) while sp does not ($0.15$--$0.48$); sp nonetheless retains classification accuracy of $0.65$--$0.94$, so the single pass keeps query dependence that iterative denoising erases.}
\label{fig:gen-sp}
\end{figure}

\paragraph{Generation--single-pass (gen--sp) correspondence.}
\textbf{The two channels agree far more often than not, and the one systematic exception is itself diagnostic.}
Across the same 1{,}502 paired conditions, $|\texttt{pos\_rate}^{\mathrm{gen}} - \texttt{pos\_rate}^{\mathrm{sp}}| < 0.10$ in 88\% of conditions and $< 0.30$ in 95\%.
Figure~\ref{fig:gen-sp}(a) shows the typical case: on safe/unsafe under alternating, LLaDA-Base's generation tracks the single pass point by point (maximum difference $0.13$), including the parity-modulated reversal at even $K \geq 24$.
The one systematic exception is LLaDA-Instruct on true/false (Figures~\ref{fig:gen-sp}(b) and~\ref{fig:gen-tails}): under alternating, generation collapses to ``true'' (\texttt{pos\_rate} $0.69$--$0.99$) at every tested $K$, while the single pass retains query dependence (classification accuracy $0.65$--$0.94$, above chance at every $K$) with only a mild ``false'' lean in its mean separation ($-1.9$ to $+0.03$).
We read this as iterative denoising amplifying the direction that a single forward pass slightly opposes.
Under grouped and blocked the gap persists, and at large $K$ under grouped the two channels exchange roles: the single pass itself slides toward ``false'' while generation stays nearer balance.
No AR model shows a comparable sustained pattern of pairing a discriminative single pass with a collapsed generation: the AR reversals that do occur (e.g., Llama-3 positive/negative under grouped) involve a single pass that has itself lost query dependence.

\section{Training-Lineage and SFT Effects}
\label{sec:sft}

Training can move label bias in opposite directions, and the two comparisons available to us isolate different causes.
\emph{Same-lineage SFT}: LLaDA-Base and LLaDA-Instruct share pre-training weights, so their differences isolate the effect of chat-style SFT (chat SFT for short).
\emph{Cross-lineage}: iLLaDA is pre-trained from scratch on a different corpus (12T tokens vs.\ LLaDA's 2.3T)~\citep{illada}, so iLLaDA-vs-LLaDA differences confound pre-training data, scale, and procedure, and we attribute them to the lineage as a whole rather than to any one factor.
We examine two effects in turn: repair across lineages, and the SFT dependence of the gen--sp decoupling; a third effect is documented in Appendix~\ref{app:neutral}.

\paragraph{Repair across lineages.}
\textbf{The positive/negative repair is a pre-training effect, and chat SFT does not reproduce it.}
On positive/negative, iLLaDA is the only healthy model while LLaDA-Base collapses toward ``positive'' (accuracy $\approx 0.51$; Section~\ref{sec:phenomenon}).
Both are pre-trained base models, so this repair reflects iLLaDA's pre-training corpus rather than fine-tuning.
The contrast extends to the antonym pairs, where it is sharper still (Section~\ref{sec:phenomenon}).
Chat SFT does not reproduce this repair: LLaDA-Instruct collapses toward ``positive'' on positive/negative under alternating at every $K \geq 16$ and under grouped by $K{=}16$--$30$, and collapses on 7 of the 10 antonym pairs.

Chat SFT also introduces a third effect: LLaDA-Instruct answers outside the label pair with a single stable third word (``neutral'') that demonstrations gradually retire (Appendix~\ref{app:neutral}).

\paragraph{gen--sp decoupling.}
\textbf{Only the chat-tuned dLLM shows a persistent decoupling between its two evaluation channels.}
The decoupling documented in Section~\ref{sec:mechanism} is itself SFT-dependent: LLaDA-Instruct collapses a still-discriminative single-pass distribution on true/false, while LLaDA-Base shows only a transient low-$K$ version of the same phenomenon (generation at $0.04$--$0.20$ at $K{=}5$--$7$ while its single pass stays $0.89$--$0.91$ accurate) that vanishes from $K{=}8$ onward.
Chat SFT thus interacts with the diffusion decoder, a reading consistent with findings that SFT can degrade context-awareness in AR models~\citep{wang2024context}.

Taken together, training effects are bidirectional and pair-dependent: chat SFT adds failure modes (stable third-word answers, direction reversal, gen--sp decoupling) without repairing the positive/negative collapse, while a different pre-training procedure repairs positive/negative and the antonym pairs entirely but leaves safe/unsafe hard-collapsed, where LLaDA-Base retains partial function.
Which label pairs are dangerous is therefore a property of the training procedure rather than of the architecture, within the models and scale tested here.
\section{Arrangement Effects}
\label{sec:arrangement}
This section asks whether rearranging the same class-balanced demonstrations can prevent collapse.
We find that arrangement sensitivity is a model$\times$pair interaction: no arrangement is uniformly safest, strong-bias collapse cannot be averted by any arrangement, and on weak-bias pairs arrangement can reverse \emph{which} label the model collapses to.

\paragraph{Arrangement sensitivity is pair and model dependent.}
\textbf{No arrangement is uniformly safest or most harmful.}
Table~\ref{tab:arrangement} and Figure~\ref{fig:arrangement} (both in the appendix) show the $K{=}8$ effect.
Averaging $|\texttt{pos\_rate} - 0.5|$ over all pairs and $K \geq 4$ shows that grouped is more severe than alternating for LLaDA-Base ($0.43$ vs.\ $0.35$), iLLaDA ($0.35$ vs.\ $0.25$), and Qwen3.5 ($0.33$ vs.\ $0.28$), while alternating is slightly more severe for Llama-3 ($0.36$ vs.\ $0.32$) and LLaDA-Instruct ($0.37$ vs.\ $0.33$).
At the individual-pair level both directions occur: iLLaDA clean/toxic at $K{=}8$ is healthy under alternating ($0.43$) but collapses under grouped ($0.01$), blocked ($0.20$), and shuffled ($0.18$); conversely, on positive/negative, collapse occurs under alternating in 77 of 111 conditions vs.\ 21 of 41 under grouped (five models pooled).
We propose to read this variation itself as diagnostic: strong variation across arrangements implicates biases introduced while processing demonstrations, whereas arrangement-invariant collapse (safe/unsafe for iLLaDA) implicates a bias read out at the output projection regardless of demonstration ordering.

\paragraph{Arrangement cannot avert strong-bias collapse.}
\textbf{No arrangement averts strong-bias collapse (Table~\ref{tab:arrangement}).}
As shown there, safe/unsafe stays pinned to ``unsafe'' for iLLaDA and Qwen3.5, and circle/square to ``circle'', under all four arrangements at every tested $K$.
Ancient/modern is likewise dangerous under every arrangement, but unlike circle/square, its collapse direction reverses with arrangement, and under blocked the two dLLM base models even return to balance (next paragraph).
The remaining models sit on a severity spectrum (Section~\ref{sec:phenomenon}): LLaDA-Base collapses under grouped (13 of 13 conditions) while merely drifting under blocked and shuffled; Llama-3 and LLaDA-Instruct retain partial function under all arrangements.

\paragraph{Arrangement-induced direction reversal.}
\textbf{The same class-balanced demonstrations can produce opposite collapsed labels depending only on their order.}
(i)~On ancient/modern, the collapse direction flips for every model: under alternating, LLaDA-Base, LLaDA-Instruct, and Llama-3 collapse to ``ancient'' ($0.93$--$1.00$) while iLLaDA drifts toward it ($0.65$; Qwen3.5 stays balanced at $0.55$), and under grouped all five collapse to ``modern'' ($0.07$--$0.17$; under shuffled, tested only for the two AR models, both also collapse to ``modern'').
Under blocked the field splits: the two dLLM base models return to balance ($0.47$ and $0.51$ at $K{=}12$), LLaDA-Instruct stays with ``ancient'' ($0.86$ at $K{=}12$), and the two AR models drift toward ``modern'' ($0.33$ and $0.37$ at $K{=}8$).
(ii)~For LLaDA-Instruct on cat/dog, the arrangement decides not just whether the model collapses but \emph{which label it collapses to}: under alternating the model collapses to ``cat'' (\texttt{pos\_rate} $0.82$--$1.00$ across $K \in \{8,12,16,30\}$), while under grouped it collapses to ``dog'' ($0.00$--$0.17$).
Arrangement is thus a latent variable in every reported ICL accuracy: a fixed test set can yield opposite systematic answers depending only on demonstration order.
\section{Conclusion and Discussion}
\label{sec:conclusion}

Label collapse is architecturally universal but graded, and persists beyond binary tasks in a preregistered four-class extension (Appendix~\ref{app:multiclass}), with the spectrum word-set-governed and batch calibration retaining near-universal repair while contextual calibration fails.
Balanced-set accuracy cannot distinguish the failure from balanced performance; separation exposes it, and the single-pass channel isolates what iterative denoising does.
Each claim's evidential weight is summarized in Appendix~\ref{app:discussion}.
These findings offer practical guidance for evaluating and hardening ICL.
More practical implications, limitations, and future extensions are discussed in Appendices~\ref{app:discussion} and~\ref{app:future_work}.

\newpage
\bibliographystyle{unsrt}
\bibliography{references}

\newpage
\appendix
\section{Related Work}
\label{sec:related}
We review label biases and demonstration order in ICL, mechanistic accounts, and diffusion language models, closing with the interpretability tools adopted in our analysis.

\paragraph{Label Biases and Demonstration Order in ICL.}
Lu et al.~\citep{lu2022fantastically} showed that permuting the same demonstrations can swing ICL accuracy from near-random to near-optimal, establishing order sensitivity as a central concern.
Min et al.~\citep{min2022rethinking} found that randomly replacing demonstration labels barely hurts performance: demonstrations primarily convey the label space, the input distribution, and the response format.
Our results expose the flip side: while the \emph{truth} of demonstration labels may be optional, \emph{which words occupy the label space} is consequential; switching the label pair from positive/negative to circle/square induces query-independent collapse across all five architectures.
Zhao et al.~\citep{zhao2021calibrate} identified majority label bias, recency bias, and common token bias~\citep{holtzman2021surface} and proposed contextual calibration; later work measured robustness to the majority-label bias induced by skewed demonstration distributions, finding robustness boundaries that vary widely across models and tasks~\citep{gupta2023robust}, and defined a typology of vanilla, context, and domain label biases~\citep{fei2023labelbias}.
All of these methods correct the label log-odds (the quantity we measure as separation) against a chosen reference point; Zhou et al.~\citep{zhou2023batch} unify the family as subtracting a context bias and show that a batch-mean reference dominates.
Our use is orthogonal: rather than correcting predictions, we read the joint distribution of the shared bias and the query-dependent component, which turns the family itself into a probe of where the bias lives (Appendix~\ref{app:calib}).
The domain-label bias of Fei et al.~\citep{fei2023labelbias} (task accuracy pinned at chance by the label words themselves) anticipates the label-word sensitivity we isolate; we extend it to strictly class-balanced arrangements, across architectures, and with the bias decomposed at the output head.
Our collapse arises under strictly class-balanced arrangements, placing it outside the majority-label-bias setting; label words have been varied only jointly with wholesale template changes~\citep{zhao2021calibrate}, not isolated under a fixed template, and, to our knowledge, collapse has not been characterized as a query-independent failure across architectures.
This is the gap our paper addresses.

\paragraph{Induction Heads and ICL Mechanisms.}
Induction heads~\citep{olsson2022induction} are a hypothesized mechanism for much of in-context learning, raising the question of whether label collapse is simply an induction failure.
The evidence in Section~\ref{sec:mechanism} argues that it is not: collapse is label-word-specific rather than pattern-specific, its direction is arrangement-invariant for strong-bias pairs, and demonstration and query information remain decodable even when the model is collapsed.
Induction heads explain how models copy patterns, but not \emph{which} label token dominates when the copied signal and the output projection disagree; our separation diagnostic measures exactly this disagreement. Complementing this mechanistic line, task-vector analyses locate the demonstration-induced adaptation in low-dimensional representation shifts~\citep{dong2025taskvectors}, and geometric studies contrast prompt-induced adaptation with the mechanism that transforms base representations during instruction tuning~\citep{kirsanov2025geometry}; both operate at the representation level, whereas our diagnostic reads the label decision directly at the output projection.

\paragraph{Diffusion Language Models and ICL.}
Discrete diffusion LMs generate text by iteratively denoising a masked sequence rather than decoding left to right~\citep{lou2024sedd}: MDLM~\citep{mdlm} showed that simple masked diffusion with a simplified objective is more performant than previously thought, LLaDA~\citep{llada} scaled it to 8B parameters, Dream~\citep{dream} demonstrated competitive reasoning and planning, and iLLaDA~\citep{illada} was trained from scratch with fully bidirectional attention (12T tokens), releasing Base and SFT variants.
SFT is known to degrade pre-trained capabilities in AR models~\citep{wang2024context}; we extend this observation to dLLMs, where it can alter label bias directions between Base and Instruct variants.
Contemporaneous work has begun to characterize dLLM in-context behavior: Piskorz et al.\ show that mask tokens are not passive placeholders (where relevant demonstrations sit relative to them shapes what the model comprehends) and that performance varies with how the same in-context examples are ordered~\citep{piskorz2026masks}.
Within this emerging literature our contribution is specific: we isolate \emph{query-independent label failure} as a phenomenon, characterize it across AR and dLLM families under matched protocols, and show that dLLMs share the AR label-collapse phenotype while raising new questions about how iterative denoising interacts with label biases.

\paragraph{Probing and Logit Lens.}
Linear probing~\citep{belinkov2022probing} trains classifiers on hidden states to test what information is encoded; the logit lens~\citep{nostalgebraist2020logit} decodes intermediate layers through the output vocabulary.
We apply the logit lens to dLLMs, reading label-token logits at the masked answer position across all transformer layers
(Section~\ref{sec:mechanism}), to locate where the label bias is set; extending trained probes and denoising-step trajectories to dLLMs remains open.

\section{Extended Experimental Details}
\label{app:details}

\paragraph{Prompt template and arrangement examples.}
Each demonstration is a sentence drawn from its pair's demonstration pool (an SST-2 movie-review sentence for the sentiment-anchored pairs, an XSTest statement for the safety-anchored pairs; Section~\ref{sec:setup}) followed by its label on the next line, and the query is followed by ``Answer:'' with no trailing space.
The following concrete example shows the alternating (PNPN) arrangement at $K{=}4$, using real demonstrations from our SST-2 pool:
\begin{quote}\small
exciting and direct , with ghost imagery that shows just enough to keep us on our toes .\\
Answer: positive\\
too slow , too long and too little happens .\\
Answer: negative\\
it 's worth seeing just on the basis of the wisdom , and at times , the startling optimism , of the children .\\
Answer: positive\\
makes for a pretty unpleasant viewing experience .\\
Answer: negative\\
\emph{light years / several warp speeds / levels and levels of dilithium crystals better than star trek xi .}\\
Answer:
\end{quote}
The four arrangements differ only in the ordering of the same demonstration pool: at every $K$, each orders the same label multiset, $\lceil K/2\rceil$ positives and $\lfloor K/2\rfloor$ negatives.
Alternating prefixes the label cycle (P, N, P, N, \ldots), so its prompts are nested across $K$; grouped concatenates the two classes (P, P, N, N at $K{=}4$); blocked alternates P- and N-blocks of up to three, shortening the tail block when $K$ is not divisible by six (PPPNNNPN at $K{=}8$); shuffled permutes the multiset at random, drawn independently at each $K$ (fixed seed).
The shuffled draw is shared by the 200 queries of a condition and by all models and word pairs at the same $K$ (PNNNPNPP at $K{=}8$), but sequences at different $K$ are unrelated; unlike the alternating prefixes, shuffled does not form a nested family across $K$.
All four are exactly class-balanced at even $K$; at odd $K$ each carries one extra positive, and alternating additionally ends on a positive demonstration (Appendix~\ref{app:parity}).
An earlier variant of this template used a trailing space after the final colon (\texttt{Answer:~}); the two variants differ in the tokenization of the label position and in bias direction, and all results in this paper use the no-space variant.

\paragraph{K grids.}
Table~\ref{tab:kgrid} lists the exact $K$ values per model and arrangement.

\begin{table}[t]
\centering
\small
\caption{Complete demonstration-count grids per model and arrangement: all $K$ values actually evaluated, unioned over word pairs. The cross-arrangement comparisons in the text use the common balanced subset $K \in \{4, 8, 12, 16, 30\}$.}
\label{tab:kgrid}
\begin{tabular}{lcccc}
\toprule
Model & PNPN & PPNN & PPPNNN & shuffled \\
\midrule
LLaDA-8B-Base & 0--30 (all) & 4--14 (all), 16, 30 & 6, 8, 9, 12, 15, 16, 18, 30 & 8, 16, 30 \\
iLLaDA-8B-Base & 0--30 (all) & 4--14 (all), 16, 30 & 6, 8, 9, 12, 15, 16, 18, 30 & 8, 16, 30 \\
LLaDA-8B-Instruct & 0--13, 15, 16, 20, 24, 30 & 4, 8, 12, 16, 30 & 8, 12, 30 & 8 \\
Llama-3-8B-Instruct & 0--30 (all) & 4, 8, 12, 16, 30 & 8, 12, 16, 30 & 8, 16, 30 \\
Qwen3.5-9B & 0--13, 15, 16, 20, 24, 30 & 4, 8, 12, 16, 30 & 8, 12, 30 & 8 \\
\bottomrule
\end{tabular}
\end{table}

\paragraph{Condition counts.}
The main grid comprises 3{,}292 conditions: 1{,}612 generation and 1{,}680 single-pass, each with 200 queries.
The 3{,}292 records include 63 generation and 81 single-pass duplicate evaluations of already-covered conditions; at the level of distinct conditions, 1{,}549 settings have both channels (every generation condition has a single-pass counterpart, and the single-pass grid additionally covers the 50 zero-shot antonym conditions of Section~\ref{sec:setup} without generation).
The analyses of Section~\ref{sec:mechanism} use the 1{,}502 of these that remain after excluding the 47 zero-shot generation conditions whose outputs are essentially unparseable (invalid rate $\geq 0.98$; the 47/50 reported in Appendix~\ref{app:calib}).
The three post-hoc equal-length pairs of Appendix~\ref{app:newpairs} (420 single-pass conditions, 405 with $K \geq 4$, and 405 generation conditions) are likewise evaluated outside the main grid and excluded from all counts above.
A small set of exploratory re-runs, consisting of 12 generation and 12 single-pass conditions using demonstration pools from other domains (fact statements for true/false, math solutions for correct/wrong, at $K \in \{4, 8\}$ under alternating), and 39 logit-lens conditions are excluded from all counts and analyses.

\paragraph{Model-specific configurations.}
LLaDA-8B-Base/iLLaDA use \texttt{transformers==4.46.3}; Qwen3.5 uses \texttt{transformers>=5.15}. All models use \texttt{torch\_dtype=torch.bfloat16}. Exact environment pins are recorded in the released code.
\paragraph{Compute resources.}
All experiments ran on a single rented NVIDIA A100-40GB GPU (bf16, inference only, no training). Total usage across all experimental versions is estimated at roughly 1{,}000 GPU-hours ($\pm$30\%), based on output-file timestamp ranges over the 67-day project period.
\paragraph{Asset licenses.}
Licenses, verified via the Hugging Face API: LLaDA-8B weights under MIT; iLLaDA-8B weights under Apache-2.0; Llama-3 (accessed via the NousResearch mirror of Meta-Llama-3-8B-Instruct) under the Meta Llama 3 Community License; Qwen3.5-9B-Base under Apache-2.0; XSTest under CC-BY-4.0; SST-2 carries no unified license (GLUE defers to the terms of the original dataset, which we cite).

\paragraph{Decoding parameters.}
For dLLMs, \texttt{block\_length=3} constrains remasking to contiguous blocks of three generated tokens: with \texttt{gen\_length=3}, the entire answer (three positions) forms a single block, refined over \texttt{steps=8} denoising iterations with low-confidence remasking. In the released code, the single-pass (sp) mode is named \texttt{ppl} for historical reasons.

\begin{figure}[t]
  \centering
  \includegraphics[width=\textwidth]{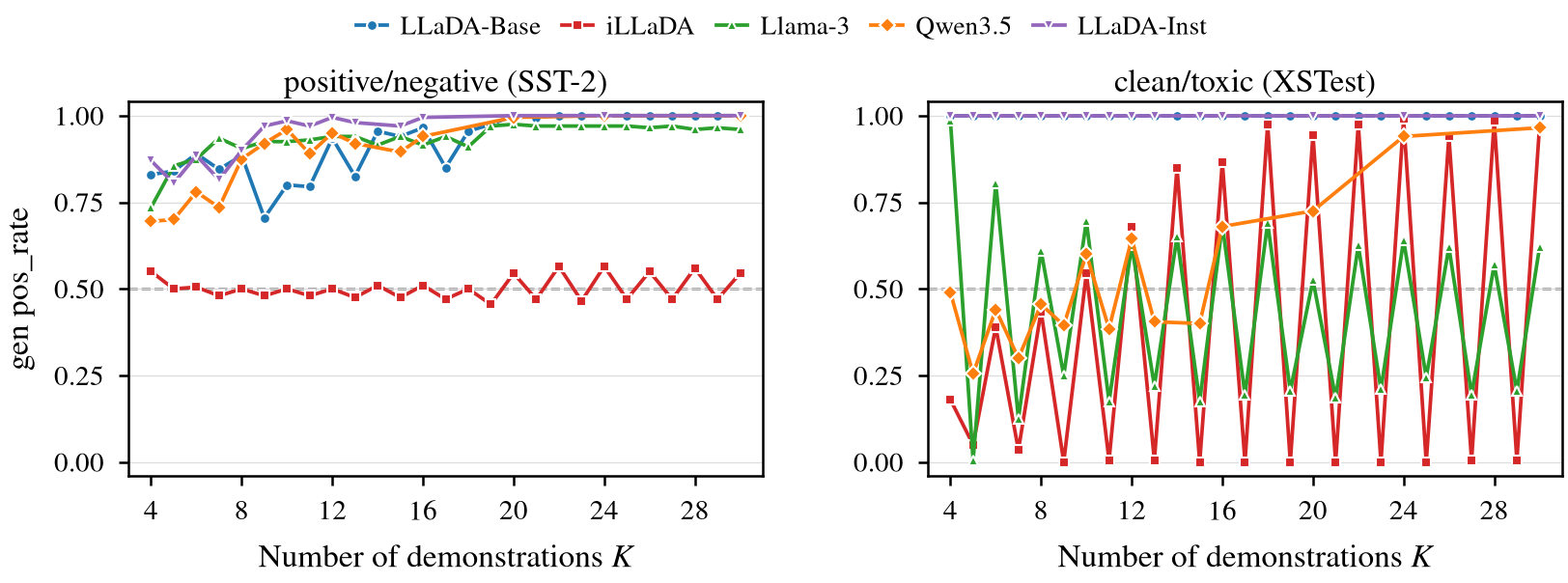}
  \caption{K-sweep generation \texttt{pos\_rate} for the remaining two representative pairs (positive/negative, clean/toxic; alternating arrangement; gray dashed line = 0.5). Complements Figure~\ref{fig:ksweep-orig4}: on positive/negative the other four models lean toward ``positive'' while iLLaDA stays balanced; on clean/toxic LLaDA-Base and LLaDA-Instruct collapse toward ``clean'' while iLLaDA and Qwen3.5 stay near balance.}
  \label{fig:ksweep-app}
\end{figure}

The single-pass criterion $\lvert\mathrm{sep}\rvert > 3$ (Section~\ref{sec:mechanism}) is used only as the model-internal predictor of the behavioral criterion, not as a severity definition.

\section{Parity Structure of the Alternating Arrangement}
\label{app:parity}

\begin{figure}[h]
  \centering
  \includegraphics[width=0.98\textwidth]{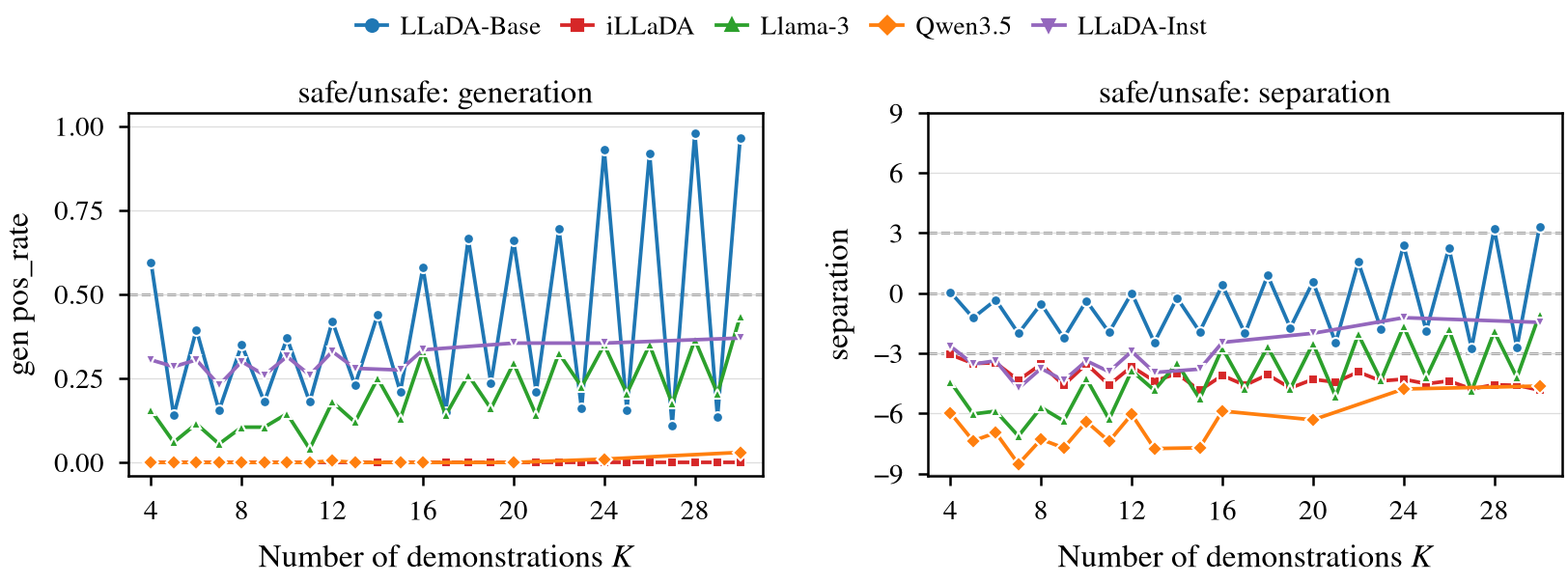}
  \caption{safe/unsafe under alternating, per model, as a function of $K$: generation \texttt{pos\_rate} (left) and single-pass separation (right).
  The two panels track each other closely (the parity oscillation of LLaDA-Base is visible in both), supporting the use of separation as a predictor of generation in this regime.
  The shaded band marks the $\lvert\mathrm{sep}\rvert \leq 3$ criterion. LLaDA-Instruct and Qwen3.5 were evaluated at the sparser $K$ grid of Section~\ref{sec:setup}.}
  \label{fig:parity-sweep}
\end{figure}

The alternating arrangement has a structural property that interacts with the
demonstration count $K$: an alternating block of even length ends on a negative
demonstration and is exactly class-balanced, whereas a block of odd
length ends on a positive demonstration and carries one extra positive.
Parity of $K$ therefore changes both the terminal label and the class
balance of the prompt. Under alternating we observe parity-structured behavior
on several model--pair combinations: the recovery trajectory of Llama-3
on safe/unsafe advances mainly on even $K$, the reversal of LLaDA-Base
on safe/unsafe appears only on even $K \geq 24$, the separation of
Llama-3 on true/false oscillates between even-$K$ values of $+3.0$ to
$+4.0$ and odd-$K$ values near zero, and iLLaDA on clean/toxic stays
near hard collapse on odd $K$ (\texttt{pos\_rate} $\leq 0.05$) while rising
through $0.18$--$0.99$ on even $K$. The direction of these oscillations
is consistently opposed to what recency bias and majority label bias
predict (the model favors the label demonstrated at the prompt's end
less, not more), and no parity structure appears in the continuous
$K{=}5$--$14$ grouped sweeps. We therefore treat alternating parity as an
arrangement--count interaction to be characterized, not as a collapse
criterion; Figure~\ref{fig:parity-sweep} shows the safe/unsafe example.

\begin{figure}[t]
  \centering
  \includegraphics[width=0.9\textwidth]{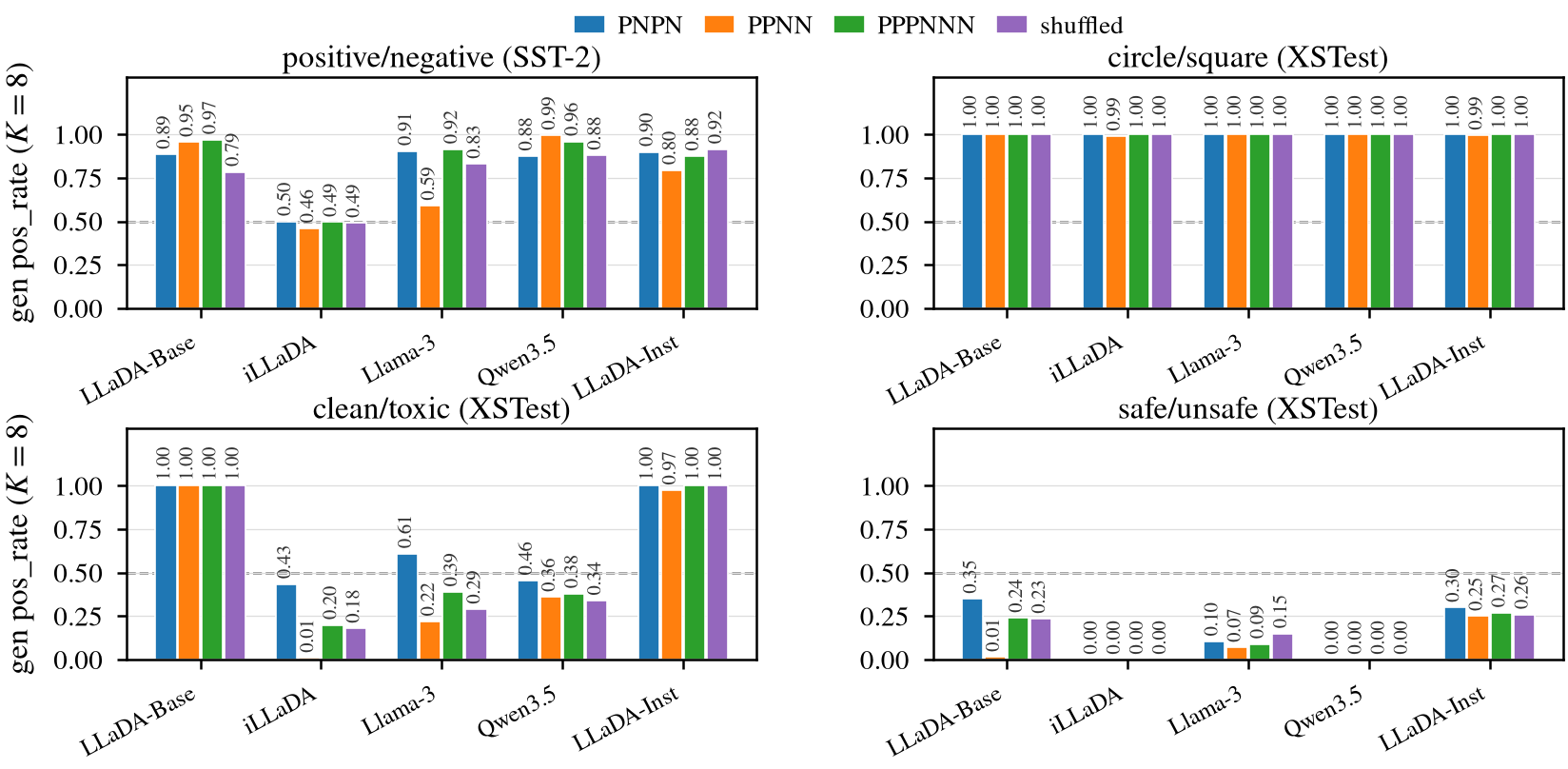}
  \caption{Arrangement comparison at $K{=}8$ (generation \texttt{pos\_rate};
  four representative pairs, five models). circle/square collapses under
  \emph{all four} arrangements for all models; safe/unsafe collapses for
  iLLaDA and Qwen3.5 under all arrangements, while Llama-3 and LLaDA-Instruct
  show arrangement-dependent severity and LLaDA-Instruct cat/dog reverses
  direction between alternating and grouped. Complements
  Table~\ref{tab:arrangement} with the four representative pairs of
  Figures~\ref{fig:ksweep-orig4} and~\ref{fig:ksweep-app}.}
  \label{fig:arrangement}
\end{figure}

\begin{table}[t]
\centering
\small
\caption{Cross-arrangement generation \texttt{pos\_rate} at $K{=}8$ (SST-2 demonstrations for cat/dog, XSTest for safe/unsafe and true/false); column heads follow the arrangement codes of Section~\ref{sec:setup}.
``Collapsed'' cells ($|\text{\texttt{pos\_rate}}-0.5|\geq0.3$) are marked \ding{55}. $^{\dagger}$Llama-3 cat/dog under blocked was not run at $K{=}8$;
the value shown is at $K{=}12$ ($0.57$; $K{=}30$ gives $0.59$).
$\triangle$ marks drift ($0.15\leq|\text{\texttt{pos\_rate}}-0.5|<0.3$).}
\label{tab:arrangement}
\begin{tabular}{llcccp{3.2cm}}
\toprule
Model & Word pair & PNPN & PPNN & PPPNNN & Pattern \\
\midrule
LLaDA-Base & safe/unsafe & 0.35\,$\triangle$ & 0.02\,\ding{55} & 0.24\,$\triangle$ & Grouped hardest \\
iLLaDA-Base & safe/unsafe & 0.00\,\ding{55} & 0.00\,\ding{55} & 0.00\,\ding{55} & All collapse \\
LLaDA-Inst. & true/false & 0.96\,\ding{55} & 0.72\,$\triangle$ & 0.83\,\ding{55} & Always ``true''-biased \\
LLaDA-Inst. & cat/dog & 0.82\,\ding{55} & 0.17\,\ding{55} & 0.47 & Direction reversal \\
Llama-3 & cat/dog & 1.00\,\ding{55} & 0.89\,\ding{55} & 0.58$^{\dagger}$ & Alternating collapse \\
Qwen3.5-9B & safe/unsafe & 0.00\,\ding{55} & 0.00\,\ding{55} & 0.00\,\ding{55} & All collapse \\
\bottomrule
\end{tabular}
\end{table}

Each spot-check $K$ of the shuffled arrangement uses an independently drawn random permutation, so its spot-check values constitute independent permutation draws: across the twelve cells with multi-$K$ shuffled coverage ($K \in \{8, 16, 30\}$), the collapse or lean direction is identical in every draw, so the shuffled readings are not specific to a single permutation. A dedicated resampling check reinforces this: ten fresh permutations per cell at $K{=}8$ on six cells spanning saturated collapse and partial lean (both model families) reproduce the original direction in all 60 draws, with per-cell standard deviation at most $0.07$.

\begin{figure}[t]
\centering
\includegraphics[width=0.62\textwidth]{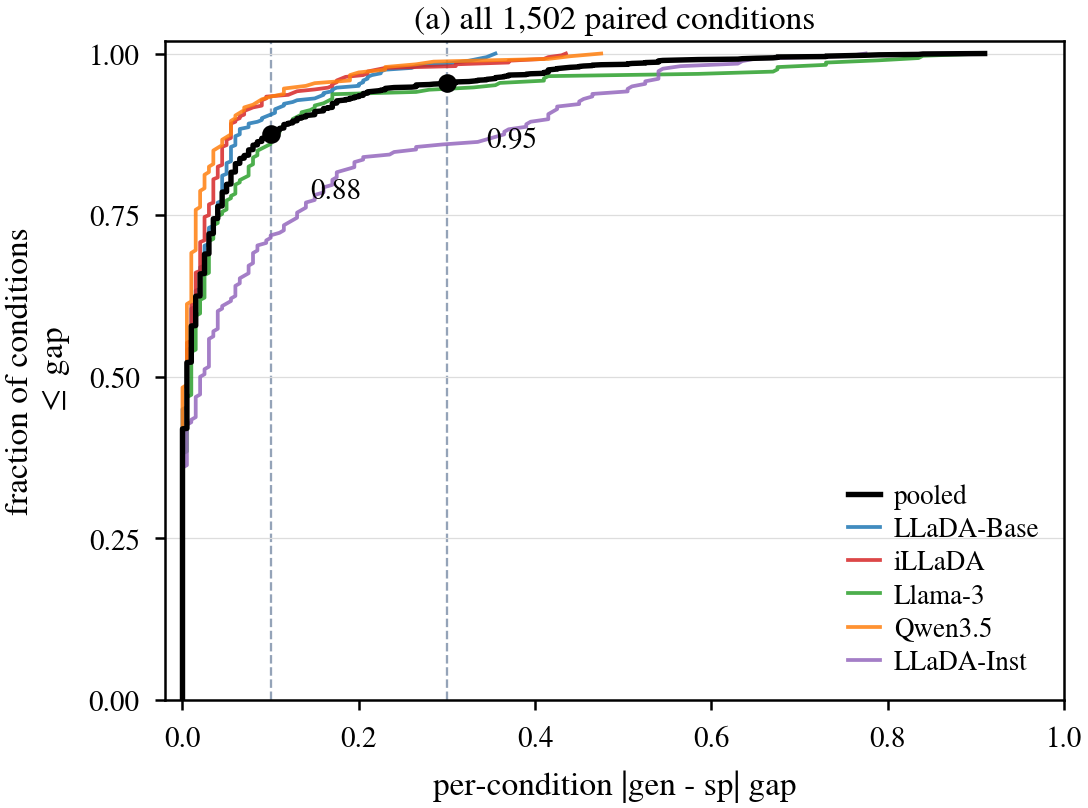}
\caption{Distribution of per-condition gen--sp gaps. Empirical CDF of the $|$gen $-$ sp$|$ \texttt{pos\_rate} gap over all 1{,}502 paired conditions (pairing of Appendix~\ref{app:details}; generation readings use the re-parsed values of Appendix~\ref{app:bpe}); thin curves per model, thick black pooled. 88\% of conditions lie below $0.10$ and 95\% below $0.30$ (dashed guides, pooled fractions marked). The tail concentrates in LLaDA-Instruct: 36 of its 256 conditions lie at or above $0.30$, versus 3--16 for each other model, and its curve is the lowest at the $0.30$ guide. Figure~\ref{fig:gen-tails} traces one cell of each kind.}
\label{fig:gen-census}
\end{figure}

\begin{figure}[t]
\centering
\includegraphics[width=\textwidth]{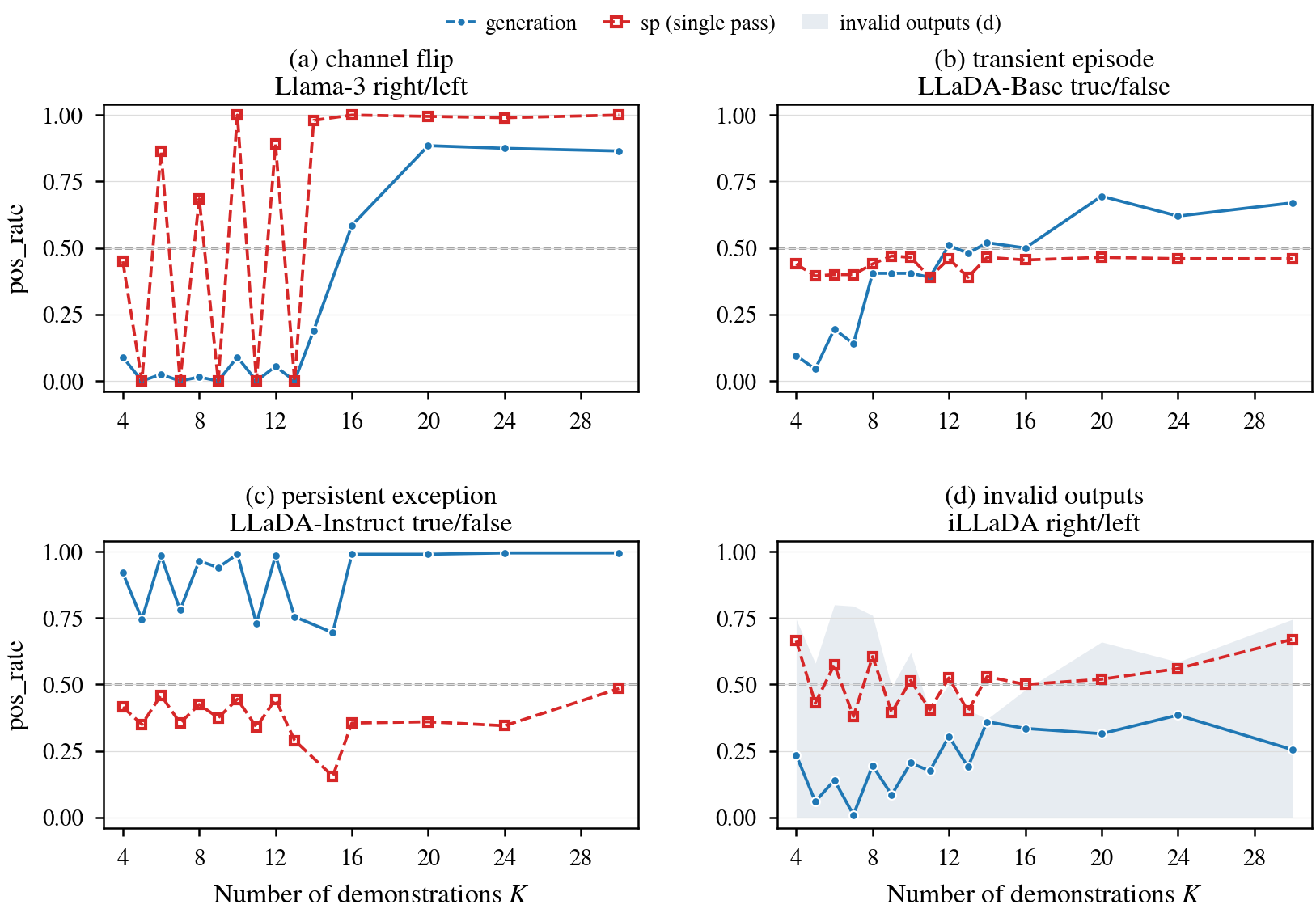}
\caption{$K$-trajectories of one alternating-arrangement cell per kind of above-threshold gen--sp behavior (blue solid: generation; red dashed: single pass; gray dashed line: balance; gray band in (d): outputs matching neither label word). \emph{(a)}~Channel flip (Llama-3 right/left): at even $K$ the two channels point to opposite labels (gap up to $0.91$) while the single pass sits at chance accuracy ($0.50$), having itself lost query information. \emph{(b)}~Transient episode (LLaDA-Base true/false): generation dips to $0.04$--$0.20$ at $K{=}5$--$7$ while the single pass stays accurate ($0.89$--$0.91$), rejoining it from $K{=}8$ onward. \emph{(c)}~Persistent exception (LLaDA-Instruct true/false): generation at $0.69$--$0.99$ at every $K$ while the single pass ($0.65$--$0.94$ accurate) stays mid-range. \emph{(d)}~Invalid outputs (iLLaDA right/left): the gray band marks outputs matching neither label word (37--80\% of queries); among the parseable remainder, generation accuracy is $0.90$--$1.00$, so its low \texttt{pos\_rate} reflects off-label emission rather than a query-blind label choice.}
\label{fig:gen-tails}
\end{figure}

The gen--sp correspondence of Section~\ref{sec:mechanism} extends beyond its two spotlighted cells: Figure~\ref{fig:gen-census} shows the distribution of per-condition gen--sp gaps over all 1{,}502 paired conditions, and Figure~\ref{fig:gen-tails} traces the $K$-sweep of one cell per kind of above-threshold behavior, making the taxonomy directly visible.

\section{Parsing Protocol and Artifact Audit}
\label{app:bpe}

Generation outputs are read at the text level: the first content token of the output (skipping leading newlines and whitespace) is matched, case-insensitively, against the two label words of the pair; \texttt{pos\_rate} is the fraction of the 200 queries whose output matches the positive label, and outputs matching neither word count as invalid.
Three artifact classes initially confounded this reading, and Table~\ref{tab:tier-map} flags any cell where more than 30\% of outputs remain invalid even after correction ($^{\dagger}$):
\begin{itemize}
\item[(i)] \textbf{Tokenizer splits.} A label word may tokenize into multiple pieces, so first-token matching fails: ``ancient'' splits in all three dLLM vocabularies (``an''+``cient''), ``toxic'' in Qwen3.5 (``to''+``xic''), and ``clean'', ``kind'', ``cruel'', ``quiet'', ``rich'', and ``wet'' split in various models.
On ancient/modern under alternating, LLaDA-Base outputs ``ancient'' for 200/200 queries at $K{=}12$ while the first-token parser records a parse error and \texttt{pos\_rate} $0.00$; text-level matching restores the correct reading.
\item[(ii)] \textbf{Format shifts.} Llama-3 capitalizes its answers (``TRUE''/``FALSE'' on true/false), and Qwen3.5 prefixes answers with a newline (or ``Answer:'') followed by free-form continuations.
Case-insensitive matching and newline skipping restore these.
\item[(iii)] \textbf{Genuine out-of-pair answers.} These are not artifacts: Qwen3.5 answers ``wrong'' or ``middle'' on right/left at small $K$ (Section~\ref{sec:phenomenon}), and such outputs remain invalid by design.
\end{itemize}
We re-parsed all 1{,}624 recorded generation conditions (the 1{,}612 main grid plus the 12 exploratory) under this protocol; per condition we retain whichever reading (original or corrected) yields the lower invalid rate.
We hand-verified three anchors: LLaDA-Base ancient/modern, alternating $K{=}12$ (\texttt{pos\_rate} $0.00 \rightarrow 1.00$; 200/200 outputs ``ancient''); Llama-3 true/false, alternating $K{=}30$ ($0.25 \rightarrow 0.85$); iLLaDA ancient/modern, alternating $K{=}4$ ($0.00 \rightarrow 0.91$).
Three cells remain flagged in Table~\ref{tab:tier-map} (iLLaDA right/left, Llama-3 right/left, Llama-3 wet/dry): their outputs leave the label vocabulary under both readings.

\section{Calibration Baselines}
\label{app:calib}

We evaluate two standard label-bias calibrations, on the scoring channel and, for the dLLMs, inside the decoding loop.
\emph{Contextual calibration (CC)}~\citep{zhao2021calibrate}: $W$ is the label-word probability averaged over three content-free queries (``N/A'', ``[MASK]'', and the empty string) appended after the same demonstrations, and the calibrated score is $\mathrm{sep} - \log(W_{\mathrm{pos}}/W_{\mathrm{neg}})$; in the generation channel the same correction is injected symmetrically into the two label logits before each remasking step (the injection is defined inside the diffusion remasking loop, which autoregressive decoding lacks, so the AR models are calibrated at the scoring level only).
\emph{PMI} (pointwise mutual information)~\citep{holtzman2021surface}: we use a query-conditional denominator (the separation of the same query at $K{=}0$), a stronger normalization than the domain-conditional premise of the original formulation.
Coverage: the scoring-channel calibration comprises 190 conditions ($K \in \{4,\ldots,30\}$, all five models; 86 of the 100 model--pair cells at $K{=}8$, the remainder lacking historical raw outputs), and the decoding injection comprises 96 conditions (three dLLMs; 20 word pairs at $K{=}8$ plus the $K$ sweep of safe/unsafe and true/false); ancient/modern is excluded from cross-version comparisons because its outputs vary across experimental versions (successive template and code iterations), a version-drift effect distinct from the run-to-run variance that deterministic decoding eliminates.

\paragraph{Repair rates.}
On the scoring channel at $K{=}8$ (45 collapsed, 41 healthy conditions), CC repairs 16/45 collapsed conditions (36\%): 9/38 (24\%) in the hard tier and all 7 conditions between the collapse and hard-collapse margins. The hard tier applies the hard-collapse margin of Section~\ref{sec:setup} per condition ($|\texttt{pos\_rate}-0.5| \geq 0.45$; all 38 sit at chance accuracy), and the between-margin conditions are the remaining seven ($0.30 \leq |\texttt{pos\_rate}-0.5| < 0.45$), while CC breaks 2/41 healthy ones; PMI repairs 4/45 (9\%), none in the hard tier, and breaks 5/41.
Per model (collapsed / CC-repaired / PMI-repaired): LLaDA 14/2/0, iLLaDA 6/2/1, LLaDA-Instruct 6/2/0, Llama-3 12/6/1, Qwen3.5 7/4/2.

\paragraph{A unified view, and three further calibrations.}
All methods above correct the same quantity: writing $\mathrm{sep}(x)$ for the label log-odds of query $x$ (Eq.~\ref{eq:sep}), every calibration predicts positive iff $\mathrm{sep}(x) > c$ for a reference point $c$ given by the content-free log-odds for CC, the zero-shot log-odds of the same query for our PMI variant, the in-domain random-word log-odds for domain-context calibration (DC)~\citep{fei2023labelbias}, the null-prompt log-odds for IntrCal~\citep{he2024promptcal}, and the test-batch mean $\bar{\mathrm{sep}} = \frac{1}{n}\sum_{i} \mathrm{sep}(x_i)$ for batch calibration (BC)~\citep{zhou2023batch}.
For two labels the BC decision reduces to a sign test of centered separations, $\hat{y}(x) = \mathbf{1}[\mathrm{sep}(x) - \bar{\mathrm{sep}} > 0]$, the log-space formulation of Zhou et al.~\citep{zhou2023batch} with softmax denominators canceled.
On the same 45 collapsed $K{=}8$ conditions: DC repairs 16/45 (the same count as CC, one condition exchanged), IntrCal repairs the identical 16 as CC (reference points read from the same biased prompt family correct the same conditions), and BC repairs the \texttt{pos\_rate} of all 45, with accuracy restored in 36/45 (25 also at the stricter $\geq 0.89$).
The nine residual conditions carry no query information in the centered log-odds; eight of the nine lie outside every method's repair set, and the exception (Llama-3 ancient/modern, a version-drift cell excluded from cross-version comparisons) is repaired by the content-free reference but not by batch centering.
Length-normalized whole-label scoring (NCC; scoring both label words' full token sequences)~\citep{sanz2025length} was evaluated on the cells selected for it: the 18 cells whose two label words differ in token count under the model's own tokenizer (six word pairs: clean/toxic, kind/cruel, ancient/modern, quiet/loud, rich/poor, wet/dry) plus ten length-symmetric controls (positive/negative and safe/unsafe, five models each).
Of these 28 cells, 15 are collapsed in the uncalibrated single pass, and NCC repairs 0 of the 15: length normalization does not move the collapsed pole.
For the dLLMs, which have no left-to-right factorization of the label probability, the score is computed by prefix teacher forcing (label tokens $1,\ldots,i{-}1$ are filled into the answer slot and the logit of token $i$ is read at its position, one forward pass per position), so it reflects the forced-conditioning distribution of the scoring channel rather than free iterative denoising; the AR models use the standard full-sequence log-probability.
Both label words are scored under the same scheme, so the forcing shapes the absolute scores, not which label wins.
BC is transductive: it requires access to the test batch, and its authors list this batch dependence as their primary limitation and leave calibration of \emph{generative} tasks as open future work~\citep{zhou2023batch}, a question our decoding-injection results above answer directly (the readout-style correction inside the loop repairs $4/96$ conditions).
We therefore read BC as a probe of where the bias lives rather than a deployable fix.
The signature transfers out of the original word pairs: on three fresh equal-length pairs (Appendix~\ref{app:newpairs}), BC returns all 298 collapsed single-pass conditions, across all five models, to balance (accuracy restored in 207) and damages none of the healthy ones.

\paragraph{The prior share draws the treatment boundary.}
On the 38 hard-tier conditions we define the prior share as $|\log(W_{\mathrm{pos}}/W_{\mathrm{neg}})|$ divided by the mean separation recomputed from the raw single-pass outputs.
A condition counts as \emph{repaired} here if contextual calibration returns it to the balance band ($|\texttt{pos\_rate}-0.5|<0.3$ after calibration); the null-prompt baseline repairs the identical set, and the decoding channel is immune on the hard tier regardless of share, so the boundary below is drawn on the scoring channel (batch calibration is excluded from this notion: it re-centers all 45, leaving no immune conditions).
Repaired conditions have prior share $0.63$--$0.95$ (median $0.74$); immune conditions $0.06$--$0.8$ (median $0.46$).
No condition with share below $0.55$ is ever repaired ($0/18$); four of the five conditions at or above $0.8$ are repaired (the exception is Qwen3.5 circle/square, share $0.80$, which stays at $1.0$); and the $0.55$--$0.8$ band is mixed. 
The prior's direction is not discriminative (every repaired and every immune condition aligns with it), so amplitude, not direction, separates the two groups.
The boundary is a property of the model--pair combination rather than of the word pair: circle/square carries prior shares of $0.95$ (Llama-3; repaired, $1.00 \rightarrow 0.49$), $0.83$ (LLaDA-Instruct; repaired to drift, $0.68$), $0.69$ and $0.61$ (LLaDA-Base, iLLaDA; stay at $1.0$), and $0.80$ (Qwen3.5; stays at $1.0$).
Repairs can also decay with $K$: LLaDA positive/negative remains repaired through $K{=}16$ ($0.56$--$0.61$) but returns to $1.0$ by $K{=}24$, as the demonstration-induced component outgrows the fixed subtraction.

\begin{figure}[t]
  \centering
  \includegraphics[width=\textwidth]{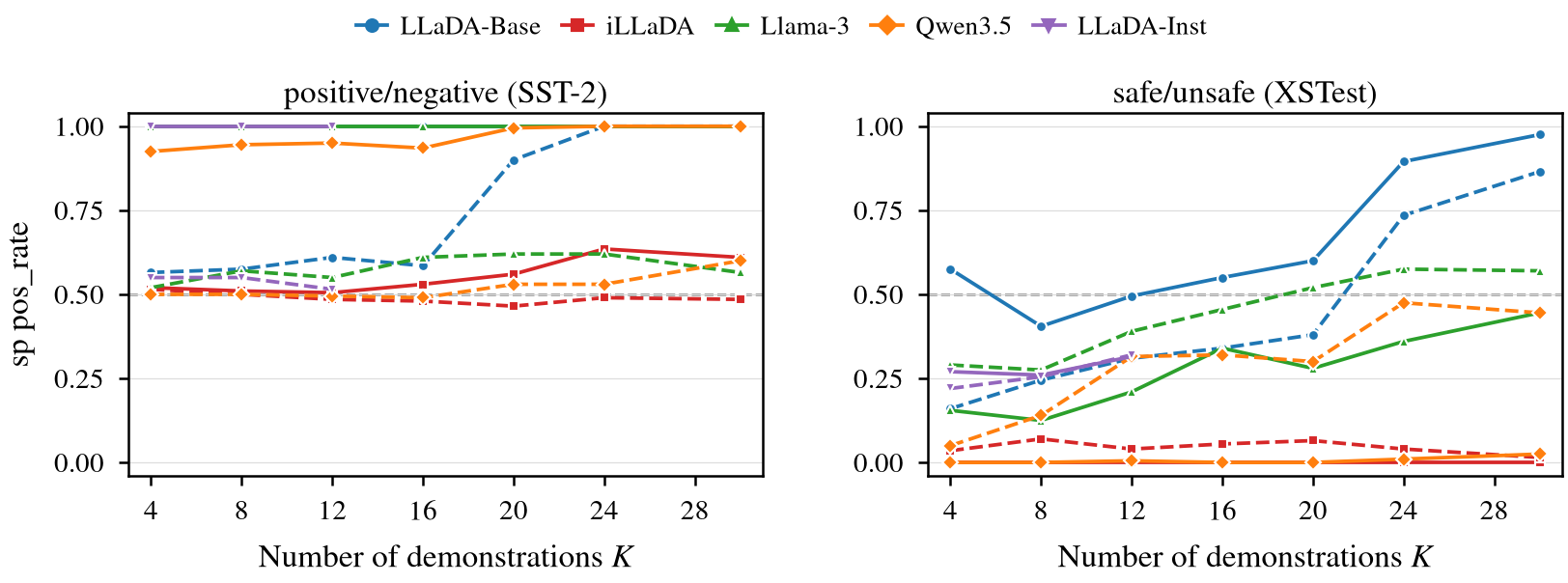}
  \caption{Single-pass \texttt{pos\_rate} before (solid) and after (dashed) contextual calibration (alternating; panels follow the word-pair convention of Figure~\ref{fig:ksweep-orig4}). Left, positive/negative: all four collapsed models move to balance ($0.49$--$0.62$; iLLaDA is balanced already, and LLaDA's repair decays beyond $K{=}20$). Right, safe/unsafe: Llama-3 is lifted steadily toward balance and Qwen3.5 at large $K$, while iLLaDA stays pinned and LLaDA overshoots at small $K$.}
  \label{fig:calib-repair}
\end{figure}

\begin{figure}[t]
  \centering
  \includegraphics[width=\textwidth]{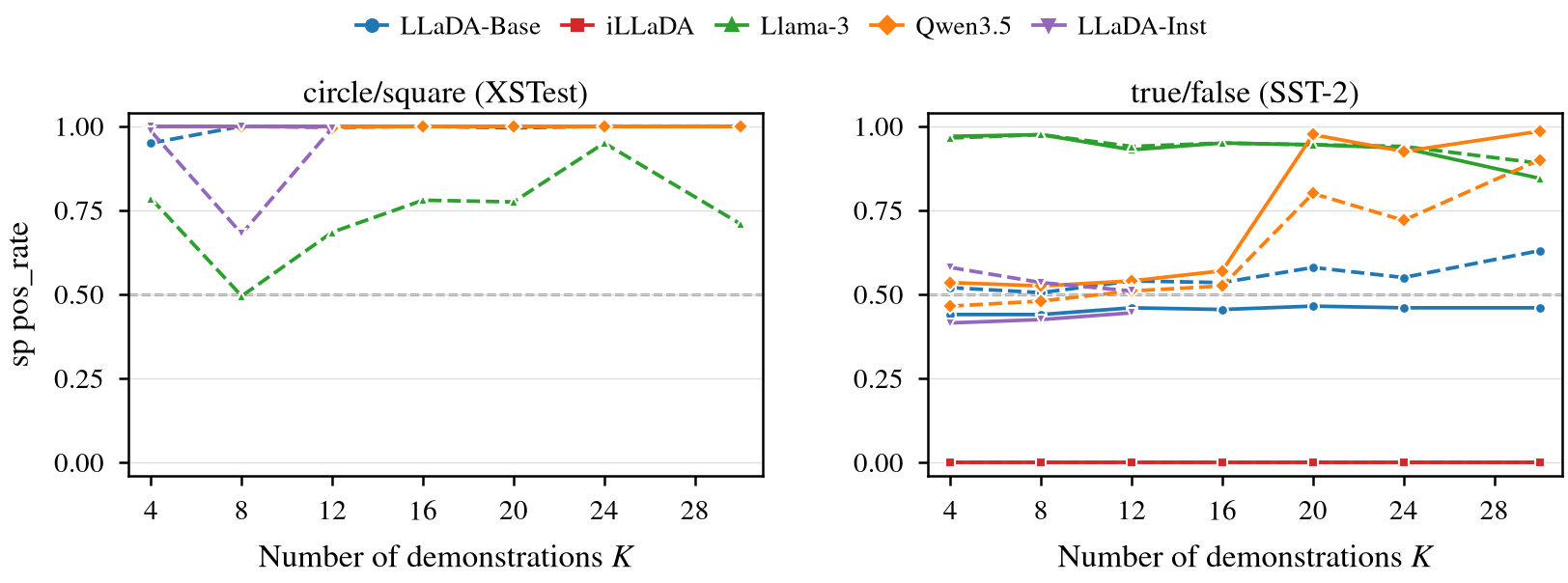}
  \caption{Word pairs that calibration barely moves (single-pass \texttt{pos\_rate}, alternating; solid before, dashed after). Left, circle/square: LLaDA, iLLaDA, and Qwen3.5 stay at $1.0$ at every $K$; Llama-3, the boundary case, recovers only at part of the grid ($0.49$ at $K{=}8$). Right, true/false: no collapsed model is repaired (iLLaDA stays at $0.0$, Llama-3 stays above $0.9$); the other three models sit near balance already.}
  \label{fig:calib-immune}
\end{figure}

\paragraph{Scoring-channel sweeps.}
Figure~\ref{fig:calib-repair} shows the word pairs that respond (solid before, dashed after).
On positive/negative, all four collapsed models move to balance ($0.49$--$0.62$; LLaDA's repair decays beyond $K{=}20$), and iLLaDA is balanced throughout.
On safe/unsafe, Llama-3 is lifted steadily toward balance ($0.12$--$0.45$ to $0.28$--$0.57$) and Qwen3.5 at large $K$ ($0.01$--$0.03$ up to $0.45$--$0.47$), while iLLaDA stays pinned and LLaDA overshoots at small $K$.
Figure~\ref{fig:calib-immune} shows the word pairs that barely move: circle/square stays at $1.0$ for LLaDA, iLLaDA, and Qwen3.5 at every $K$ (Llama-3 recovers only at part of the grid), and on true/false no collapsed model is repaired (iLLaDA stays at $0.0$, Llama-3 above $0.9$).

\begin{figure}[t]
  \centering
  \includegraphics[width=\textwidth]{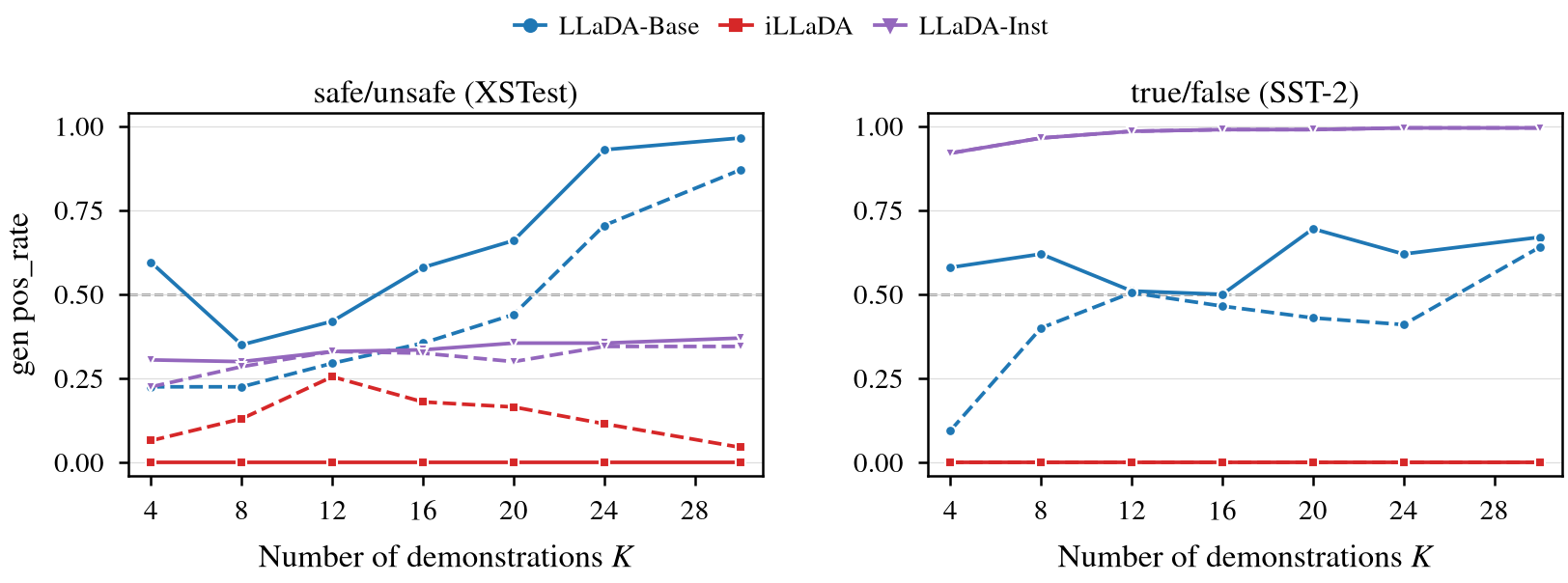}
  \caption{Generation \texttt{pos\_rate} before (solid) and after (dashed) injecting the contextual-calibration correction into dLLM decoding (alternating; three dLLMs, word-pair panels as in Figure~\ref{fig:ksweep-orig4}). Left, safe/unsafe: the correction shifts LLaDA-Base partway toward balance (overshooting at $K{=}4$) and lifts iLLaDA transiently off the floor ($0.255$ peak at $K{=}12$). Right, true/false: decoding is essentially unchanged. The injection repairs margins, not the hard tier.}
  \label{fig:calib-gen}
\end{figure}

\paragraph{Decoding-channel injection.}
Figure~\ref{fig:calib-gen} shows the injected decoding sweep for the two anchor word pairs (three dLLMs; solid before, dashed after injection).
Four conditions repair cleanly: LLaDA positive/negative ($0.885 \rightarrow 0.510$), LLaDA-Instruct positive/negative ($0.900 \rightarrow 0.500$), iLLaDA easy/hard ($0.96 \rightarrow 0.765$), and iLLaDA safe/unsafe at $K \geq 12$ (from $0.0$ to a $0.255$ peak at $K{=}12$, declining to $0.045$ at $K{=}30$).
The hard tier is immune in this channel as well: circle/square remains at $0.99$--$1.0$ for all three dLLMs, iLLaDA true/false remains at $0.0$, and on true/false the injection leaves decoding essentially unchanged throughout.
Two failure modes deserve note: overshoot, where a healthy condition is pushed past balance (LLaDA safe/unsafe $K{=}4$: $0.595 \rightarrow 0.225$), and parsing cost, where calibrated outputs leave the label vocabulary (LLaDA light/dark: invalid rate $0.01 \rightarrow 0.23$).
The scoring and decoding verdicts agree on 18/26 collapsed conditions; the eight disagreements concentrate on conditions where the correction is small relative to decoding dynamics.

\paragraph{Zero-shot prior statistics.}
Over the 45 collapsed model--pair combinations, the zero-shot separation agrees with the collapse direction in 33/45; restricting to combinations with a meaningful zero-shot prior ($|$sep$|>1$) raises this to 24/29 (83\%).
The unigram logits, defined on the 37 collapsed combinations whose two label words are single tokens, agree in 28/37, rising to 18/21 (86\%) among those with $|$logit diff$|>1$.
The two prior measurements cross-validate each other over all 82 single-token model--pair combinations, healthy or collapsed (sign agreement 73\%, $r{=}0.68$).
Zero-shot generation produces neither label in 47/50 conditions (invalid rate $\geq 0.98$): with no demonstrations almost no model enters the label vocabulary, so the readable quantity at $K{=}0$ is the prior rather than behavior.
Excluding circle/square, whose collapse trivially follows its token prior, changes the full-set agreement from 33/45 and 28/37 to 28/40 and 23/32, qualitatively unchanged.

\section{Equal-Length New Label Pairs}
\label{app:newpairs}
\begin{figure}[t]
\centering
\includegraphics[width=\textwidth]{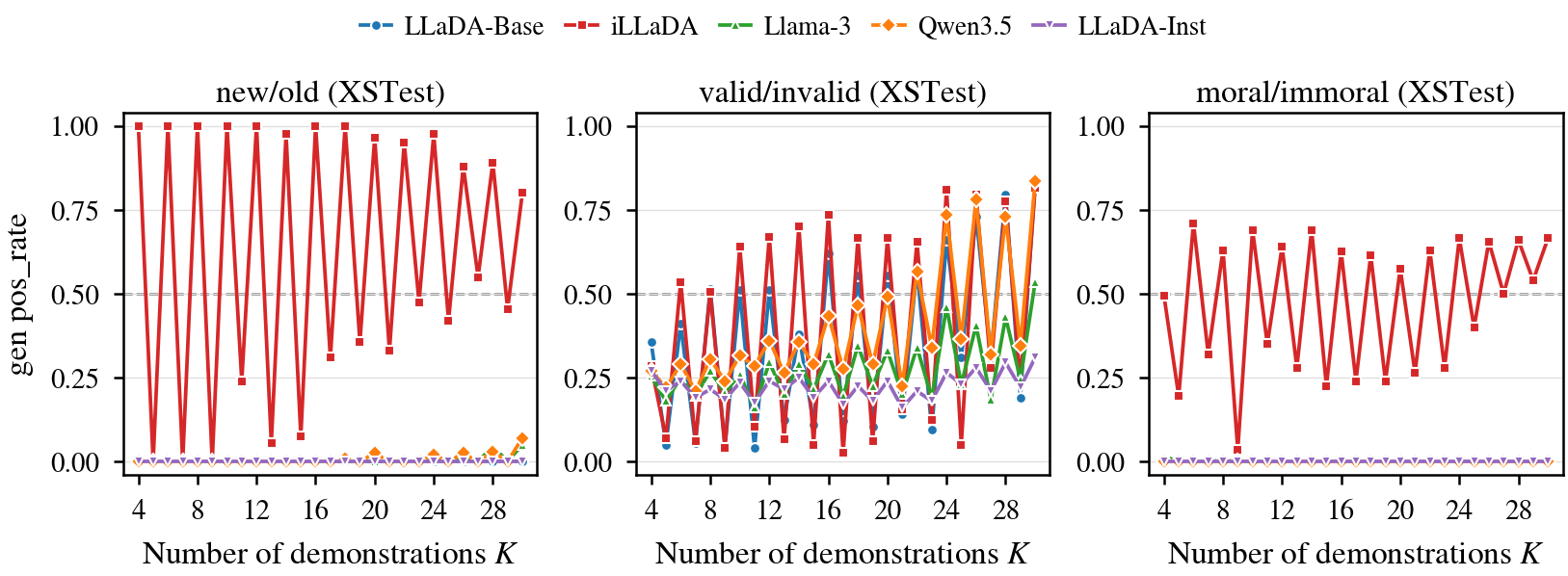}
\caption{Generation \texttt{pos\_rate} on the three equal-length pairs of Appendix~\ref{app:newpairs} (alternating arrangement, 200 queries, $K{=}4$--$30$).
\emph{Left:} new/old collapses toward ``old'' within $K{\le}8$ for LLaDA-Base, LLaDA-Instruct, Llama-3, and Qwen3.5, while iLLaDA alternates with parity throughout (even $K$ at ``new'', odd $K$ at ``old'').
\emph{Middle:} moral/immoral collapses toward ``immoral'' for four models while iLLaDA stays functional.
\emph{Right:} valid/invalid shows a parity-structured drift in LLaDA-Base, iLLaDA, and Qwen3.5 (odd $K$ toward ``invalid'' from the lowest $K$, even $K$ rising toward ``valid'' as $K$ grows, Qwen3.5 joining only at $K{\approx}20$), a mild ``invalid'' lean for LLaDA-Instruct, and near-balance for Llama-3.
The gray dashed line marks $\texttt{pos\_rate}=0.5$.}
\label{fig:newpairs}
\end{figure}

Three fresh word pairs, constructed after the analyses above and evaluated with the identical protocol (XSTest pool, alternating arrangement, $K{=}4$--$30$, 200 queries, both channels for all five models), test whether token-length asymmetry drives collapse: new/old and valid/invalid are single-token on both sides for all five tokenizers, and moral/immoral is a matched two-token pair (\texttt{m}+\texttt{oral} vs.\ \texttt{imm}+\texttt{oral}).
Collapse survives equal length: new/old collapses outright for four models (LLaDA, LLaDA-Instruct, Llama-3, Qwen3.5, all toward ``old'') while iLLaDA's generation alternates with parity over the whole range (even $K$ at ``new'', odd $K$ at ``old''); moral/immoral collapses toward ``immoral'' for four models while iLLaDA drifts mildly ($0.57$) with accuracy intact ($0.82$); valid/invalid collapses slowly rather than never: its generation medians stay balanced ($0.21$--$0.52$), but LLaDA-Base and iLLaDA are parity-structured from the lowest $K$ and Qwen3.5 joins at $K{\approx}20$, with even $K$ toward ``valid'' (up to $0.83$) and odd $K$ toward ``invalid'' (down to $0.04$), while LLaDA-Instruct leans ``invalid'' throughout and Llama-3 stays near balance (Figure~\ref{fig:newpairs}).
Equal length thus neither prevents nor guarantees collapse, and the word pair fixes not only whether but also how fast and in what pattern collapse arrives: new/old completes within $K{\le}8$ for the four pinned models (iLLaDA's alternation persists to $K{=}30$), valid/invalid builds its parity structure from the lowest $K$ and strengthens it gradually, and the alternating arrangement imprints its parity on the slow case, with even $K$ (last demonstration ``invalid'') collapsing toward ``valid'' and odd $K$ the reverse.
Which pairs are dangerous remains a property of the word pair, not of its token lengths.

The direction measurements of Section~\ref{sec:mechanism} replicate on the new pairs: among the nine collapsed model--pair combinations, the zero-shot ($K{=}0$) separation predicts the collapse direction in 7/9 and empty-context unigram logits in 7/9, the two measurements agreeing with each other in 14/15 combinations overall; the unigram direction matches the corpus-lower-frequency side in 7/9, consistent with the 7/14 rate on the original pairs.
The two zero-shot exceptions are a single phenomenon: for LLaDA-Base and LLaDA-Instruct on moral/immoral, both the zero-shot and unigram directions point to ``moral'' ($+2.05$ and $+1.04$) yet the in-context separation and generation collapse toward ``immoral'' (LLaDA-Base, $\mathrm{sep}=-8.62$ at $K{=}30$); the demonstration context can \emph{reverse} the direction inherited from the prior, so direction prediction is probabilistic rather than deterministic.
We exclude the slowly drifting valid/invalid combinations from these counts: a parity oscillation around balance has no single collapse direction, and its median flips between channels (LLaDA, $0.38$ generation vs.\ $0.54$ single-pass).

\begin{figure}[t]
\centering
\includegraphics[width=\textwidth]{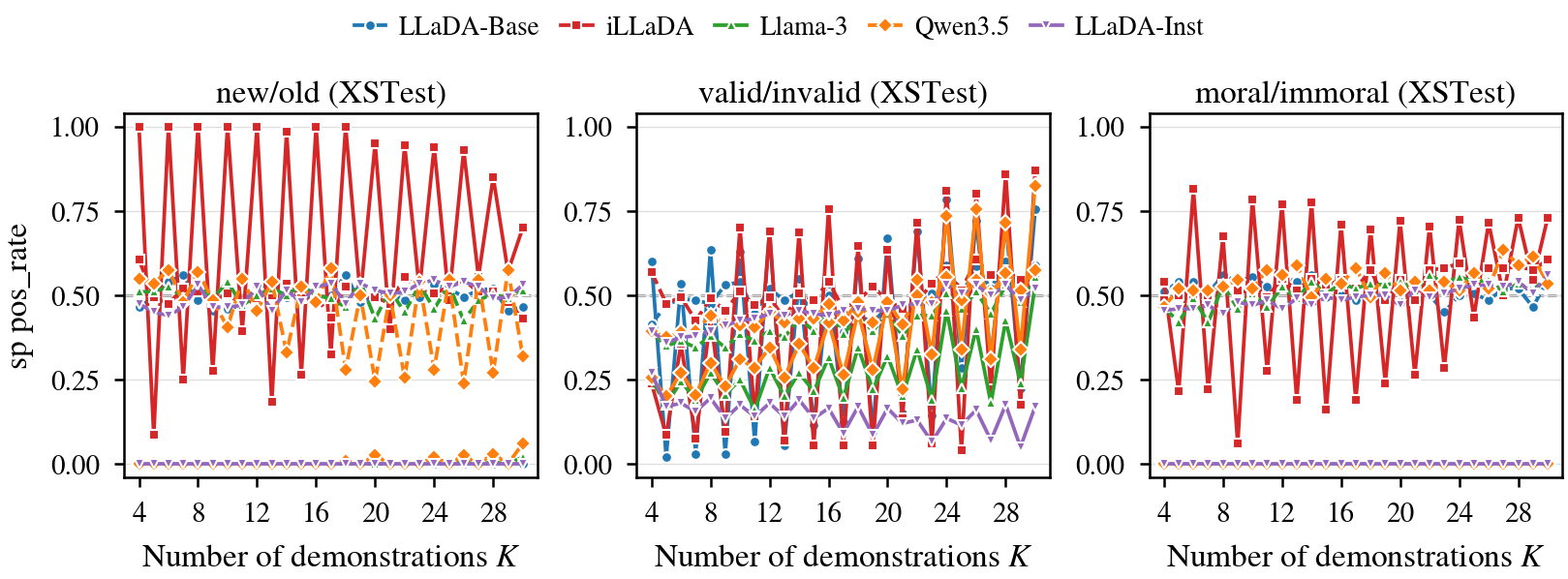}
\caption{Single-pass \texttt{pos\_rate} on the three equal-length pairs of Appendix~\ref{app:newpairs} (alternating, $K{=}4$--$30$; solid = uncalibrated, dashed = batch-calibrated, gray dashed line = balance). On new/old and moral/immoral, four models are pinned to a single label while iLLaDA's single pass alternates with the parity of $K$; on valid/invalid, LLaDA-Base and iLLaDA oscillate with parity from the lowest $K$, Qwen3.5 joins at $K{\approx}20$, Llama-3 shows the same parity at small amplitude, and LLaDA-Instruct leans mildly toward ``invalid''. Subtracting the test-batch mean of the log-odds (Appendix~\ref{app:calib}) centers all 298 collapsed conditions without disturbing the healthy ones.}
\label{fig:newpairs-bc}
\end{figure}

Calibration behaves as in Appendix~\ref{app:calib}: contextual calibration repairs only part of the collapsed conditions, and unstably (LLaDA new/old: $0.40$ at $K{=}8$, back to $0.065$ at $K{=}16$), whereas batch calibration (Figure~\ref{fig:newpairs-bc}) returns all 298 collapsed single-pass conditions (of 405 over the five models) to balance and restores accuracy ($\geq 0.7$) in 207 of them, with no damage to the healthy ones; the parity-drift conditions are included in the 298, and centering flattens their oscillation as well ($0.34$--$0.60$).
Finally, the multi-token pair reprises the parsing hazard of Appendix~\ref{app:bpe}: first-token matching misreads every moral/immoral generation (invalid rate $\approx 1.0$, the first tokens being \texttt{m}/\texttt{imm}), which text-level prefix matching restores (invalid rate $0.00$ after re-parsing).

\section{Neutral Third-Word Answers under Chat SFT}
\label{app:neutral}

Chat SFT introduces a stable third-word answer that demonstrations gradually retire.
LLaDA-Instruct sometimes answers with a word outside the label pair, and the word is almost always the same one: ``neutral''.
The small-$K$ alternating runs with $1 \leq K < 4$ (Table~\ref{tab:kgrid}), used only in this analysis, support a finer view of this effect: from $K{=}3$ through $K{=}8$, ``neutral'' accounts for 91--97\% of all out-of-pair outputs, at rates of $0.09$--$0.19$ per condition, before declining to zero by $K{=}24$ (at $K{=}2$ the share is 77\%, the remainder being truncated spellings of ``negative'').
Two regimes thus coexist in the chat-tuned model. Without demonstrations, it reaches for whatever answer words it knows: in zero-shot conditions, the out-of-pair outputs are scattered across the label vocabularies of other tasks (numbers, ``yes''/``no'', star ratings), with no ``neutral'' at all.
With even a few demonstrations, it converges on this single stable third word, which further demonstrations gradually retire.
LLaDA-Base never produces out-of-pair answers at any $K$, so the effect is specific to the chat-tuned variant.

\section{Discussion}
\label{app:discussion}
We distill the findings into practical guidance (what to measure and what to control) and state limitations.

\paragraph{Evidential status of the claims.}
The phenomena (tier structure, arrangement effects, gen--sp decoupling) are direct observations over the full grids.
The token-prior origin of collapse directions rests on converging observational measurements (zero-shot separation agreement 33/45, unigram agreement 28/37, and the corpus-frequency control at 7/14; Section~\ref{sec:mechanism}), not on intervention.
The strongest causal statement we make (that one shared per-condition offset suffices to restore balance) rests on the batch-calibration intervention (45/45 collapsed conditions re-centered at $K{=}8$; Appendix~\ref{app:calib}).
Training-lineage attributions (Section~\ref{sec:sft}) are confounded by design (Limitation~(v)) and are stated as associations.
Three implications follow from the results above.
\paragraph{Practical implications.}
\textbf{Separation is a cheap, model-agnostic pre-deployment check; which words serve as labels is a design decision; and training can move label bias in either direction.}
One forward pass at the answer slot anticipates behavioral collapse with 92.5\% precision, and its disagreements with generation mark exactly the conditions worth auditing; we recommend it as a standard ICL metric, especially for diffusion models, where a single pass is cheap compared with iterative decoding.
Label-token choice can systematically disable ICL (circle/square collapses for every architecture tested; the pair can even decide the failure mode, as on ancient/modern), so practitioners should audit label tokens with separation before deployment and report label words alongside ICL accuracy.
Because training can strengthen one label pair's bias while leaving another untouched, dangerous pairs cannot be predicted from the training data alone; training data should cover diverse label tokens, as SFT-data composition is known to shape context reliance~\citep{wang2024context}, and measuring separation across label pairs after training flags the dangerous ones before deployment.

\paragraph{Limitations.}
(i)~We evaluate 8B/9B models only; larger dLLMs may behave differently.
(ii)~The 20 word pairs, though diverse, do not cover all semantic domains, and arrangement coverage is uneven across pairs.
(iii)~Our mechanistic account stops at the output projection; which properties of the training data produce a dangerous pair remains open.
(iv)~The alternating-parity structure is characterized empirically but not explained; a controlled study is left for future work.
(v)~Cross-lineage comparisons confound pre-training data, scale, and pre-training procedure; our SFT-specific conclusions rest solely on the same-lineage LLaDA-Base/Instruct pair, and we did not evaluate iLLaDA-Instruct.
(vi)~The separation quantity itself is standard, the label log-odds that calibration methods operate on~\citep{zhao2021calibrate,zhou2023batch}; our contribution is the diagnostic protocol and the phenomena it reveals, not the quantity.
(vii)~Each pair uses one fixed demonstration pool and template; a pool-swap probe in the multi-class census (Appendix~\ref{app:multiclass}) found collapse status and direction highly consistent across independently drawn pools, but sensitivity to the choice of demonstration sentences in the binary grids remains unmeasured.

\section{Future Work} \label{app:future_work}

Several directions remain:
(i)~extending separation analysis to more dLLM architectures (MDLM; Dream variants enter the multi-class census of Appendix~\ref{app:multiclass});
(ii)~identifying which training-data properties make a label pair dangerous, and designing data accordingly;
(iii)~layer-level interventions that intercept collapse before the output projection;
(iv)~a mechanistic account of gen--sp decoupling in iterative denoising;
(v)~sweeping decoding configurations (denoising step counts, re-masking strategies) to test whether the gen--sp decoupling is intrinsic to iterative denoising or specific to the sampler configuration; the multi-class census of Appendix~\ref{app:multiclass} is a first step beyond binary label sets.


\section{Multi-Class Extension: Four-Way Classification on AG News}
\label{app:multiclass}

\paragraph{Design.}
To test whether the binary findings transfer beyond two-class tasks, we ran a preregistered census (judgment thresholds 0.70/0.625/0.55/0.40 for collapse/drift/gray-zone/healthy, fixed before data collection) on four-way AG News topic classification~\citep{zhang2015agnews}.
Seven models: the five of Section~\ref{sec:setup} plus Dream-7B Base and Instruct~\citep{dream} (not in the binary grid).
Twelve four-word label sets in four groups: \emph{task-aligned} (A1 world/sports/business/technology; A2 international/athletics/finance/science; A3 global/games/commerce/research), \emph{semantically related} (W1 war/victory/profit/discovery), \emph{task-unrelated} (U1 circle/square/triangle/star; U2 red/blue/green/yellow; U3 cat/dog/bird/fish; U4 north/south/east/west; U5 spring/summer/autumn/winter; U6 one/two/three/four), and \emph{merged binary pairs} (V1 strong/weak/fast/slow; V2 large/small/black/white).
Grid: $K \in \{4, 8, \ldots, 28\}$, three arrangements (cyclic ABCD, grouped, per-query shuffled), 200 queries per condition, 2{,}901 conditions total; the two AR models cover the A1/U1/V2 subset.
All 48 label words fit the \texttt{gen\_length=3} budget under all seven tokenizers (first tokens mutually distinct within each set), and byte-level anchors against the binary pipeline reproduce its readings exactly.
Collapse reads $\max_i \texttt{rate}(w_i) \geq 0.70$ over the four label words (generation) or the same threshold on the per-class prediction rate (single pass); out-of-set outputs are tracked separately.

\begin{table}[t]
\centering
\caption{Multi-class collapse spectrum: collapsed generation conditions (threshold $0.70$) over $K \geq 12$, main demonstration pool, by label set. Dream rows: \emph{all} cells are out-of-set collapse (out-of-set rate $1.000$), not health (text below). AR rows cover A1/U1/V2 only.}
\label{tab:multiclass-spectrum}
\scriptsize
\begin{tabular}{l cccccccccccc}
\toprule
Model & A1 & A2 & A3 & W1 & U1 & U2 & U3 & U4 & U5 & U6 & V1 & V2 \\
\midrule
LLaDA & 0/15 & 3/15 & 8/15 & 7/15 & 15/15 & 10/15 & 10/15 & \textbf{15/15} & 10/15 & 10/15 & 11/15 & 9/15 \\
LLaDA-Inst. & 8/15 & 9/15 & 9/15 & 5/15 & 10/15 & 11/15 & 10/15 & \textbf{15/15} & 6/15 & 10/15 & 14/15 & 11/15 \\
iLLaDA & 1/15 & 0/15 & 5/15 & 0/15 & 13/15 & 10/15 & 10/15 & \textbf{15/15} & 10/15 & 9/15 & 11/15 & 10/15 \\
Dream-Base & 0/15* & 0/15* & 0/15* & 0/15* & 0/15* & 0/15* & 0/15* & 0/15* & 0/15* & 0/15* & 0/15* & 0/15* \\
Dream-Inst. & 0/15* & 0/15* & 0/15* & 0/15* & 0/15* & 0/15* & 0/15* & 0/15* & 0/15* & 0/15* & 0/15* & 0/15* \\
Llama-3 & 0/15 & --- & --- & --- & 10/15 & --- & --- & --- & --- & --- & --- & 6/15 \\
Qwen3.5 & 3/15 & --- & --- & --- & 10/15 & --- & --- & --- & --- & --- & --- & 10/15 \\
\bottomrule
\end{tabular}
\end{table}

\paragraph{Collapse persists, and the label word set governs it.}
Over the task-unrelated (U) sets at $K \geq 12$, 219 of 480 generation conditions collapse (45.6\%); the merged-binary V sets behave similarly.
Which words occupy the label set, not the architecture family, organizes the spectrum (Table~\ref{tab:multiclass-spectrum}): U4 (north/south/east/west) collapses in every LLaDA-lineage condition (15/15 each, all toward ``west''), reproducing the binary north/south failure in a harder form, while W1 (war/victory/profit/discovery) barely collapses on iLLaDA and Llama-3.
The two AR models' collapse spectra agree with the five dLLMs' (mean per-condition collapse-label agreement 0.743 and 0.721 for Llama-3 and Qwen3.5) at or above the dLLM-internal mean of 0.644; the word set, not the architecture, is the primary organizing variable, now across both architecture families.
Chat-SFT can \emph{worsen} task-aligned collapse: LLaDA-Instruct collapses on 26 of 45 A-group conditions versus 11 of 45 for LLaDA-Base, echoing the binary finding that label-word effects dominate task semantics.

\paragraph{Direction is prior-influenced but no longer prior-dominated.}
The collapse word matches the $K{=}0$ single-pass argmax in 144 of 435 collapsed conditions (33.1\%, binomial $p = 9.2 \times 10^{-5}$ against the 25\% chance baseline), and the $K{=}0$ argmax is itself strongly shared across models (e.g., ``strong'' for V1 in five of seven models, ``west'' for U4 in five of seven), pointing to a token-frequency origin rather than an architectural one.
The contrast with the binary census is informative: there, three quarters of collapse directions matched the zero-shot prior; with four competing words, the prior's share drops to about a third; prior influence persists but competing word-set structure matters more as the label space grows.

\paragraph{Calibration repair splits sharply.}
On the 894 collapsed single-pass conditions, batch calibration repairs 824 (92.2\%, per model 83--98\%, versus 45/45 in the binary census), while contextual calibration repairs 49 (5.5\%, versus 16/45 binary): the content-free-reference estimate that suffices for binary word pairs fails to track the per-class biases of four-way label sets.
The transductive batch-mean reference remains near-universal.

\paragraph{Two negative results and a third failure mode.}
The hypothesized alignment gradient (task-aligned $<$ semantically related $<$ task-unrelated severity) does not hold: mean large-$K$ collapse severity is 0.348 (A), 0.248 (W), and 0.521 (U); the W group is the \emph{least} collapsed, and the A-group mean is inflated by the SFT effect above.
The gen--sp decoupling that motivated the binary single-pass analysis is nearly absent: one condition of 2{,}901 pairs a healthy single pass ($\leq 0.40$) with a collapsed generation ($\geq 0.70$), so in the four-way setting the two channels collapse together.
Finally, the Dream variants exhibit a third failure mode: at every $K \geq 12$ their generations match \emph{none} of the four label words (out-of-set rate $1.000$) while their single-pass readings are well-formed and largely collapsed (repaired by batch calibration at 98\%). The model fails by leaving the label vocabulary rather than by repeating one word of it.
A pool-swap probe (an independently drawn demonstration pool for 16 LLaDA/Dream-Base conditions) preserves collapse status in 15/16 and collapse direction in 9/10 doubly-collapsed conditions, supporting the robustness of the census to demonstration-pool choice (Limitation~(vii)).

\paragraph{Relation to the binary census.}
The extension confirms the headline phenomena (collapse is easy to trigger with the wrong label words, invisible to balanced accuracy, and repairable by a shared per-condition offset) while sharpening two boundaries: the token prior competes with word-set structure rather than dominating it, and the gen--sp decoupling observed in the binary grid does not reappear at four classes, so it should not be assumed generic to iterative denoising.


\section{Reproducibility} \label{app:reproducibility}

All experimental configurations, code, and data will be released upon acceptance. The experimental setup uses publicly available models and datasets, with full hyperparameters described in Section~\ref{sec:setup}.

\newpage
\section*{NeurIPS Paper Checklist}

\begin{enumerate}

\item {\bf Claims}
    \item[] Question: Do the main claims made in the abstract and introduction accurately reflect the paper's contributions and scope?
    \item[] Answer: \answerYes{}
    \item[] Justification: The abstract and introduction state four contributions: (i)~label collapse is architecturally universal across five AR and dLLM models, (ii)~standard accuracy conceals it, (iii)~separation is a model-agnostic diagnostic, and (iv)~arrangement can reverse collapse direction on weak-bias pairs but cannot avert strong-bias collapse. These match the experimental results in Sections~3--6, and the scope of the evidence (models, word pairs, arrangements, $K$ ranges; Section~2 and Appendix~\ref{app:details}) matches the claims; generality boundaries are discussed in Appendix~\ref{app:discussion}.

\item {\bf Limitations}
    \item[] Question: Does the paper discuss the limitations of the work performed by the authors?
    \item[] Answer: \answerYes{}
    \item[] Justification: Limitations are discussed in Appendix~\ref{app:discussion} (Discussion), covering model scale and family coverage, the restriction to English binary classification with fixed templates, the mechanism being localized to the output projection rather than intermediate layers, the unexplained parity structure of alternating arrangements, and base-versus-instruct lineage confounds.

\item {\bf Theory assumptions and proofs}
    \item[] Question: For each theoretical result, does the paper provide the full set of assumptions and a complete (and correct) proof?
    \item[] Answer: \answerNA{}
    \item[] Justification: The paper is empirical; it contains no theorems, lemmas, or formal proofs.

    \item {\bf Experimental result reproducibility}
    \item[] Question: Does the paper fully disclose all the information needed to reproduce the main experimental results of the paper to the extent that it affects the main claims and/or conclusions of the paper (regardless of whether the code and data are provided or not)?
    \item[] Answer: \answerYes{}
    \item[] Justification: Section~2 specifies models, template, arrangements, metrics, and demonstration configurations; Appendix~\ref{app:details} gives the complete $K$ grids, condition counts, and per-model decoding parameters; Appendix~\ref{app:bpe} documents the parsing protocol and artifact audit. Generation is deterministic (temperature~0) with fixed demonstration pools and seeded per-$K$ shuffles, and Appendix~\ref{app:reproducibility} states that all configurations, code, and data will be released upon acceptance.

\item {\bf Open access to data and code}
    \item[] Question: Does the paper provide open access to the data and code, with sufficient instructions to faithfully reproduce the main experimental results, as described in supplemental material?
    \item[] Answer: \answerNo{}
    \item[] Justification: Code and data are not publicly released at submission time; the paper states they will be released upon acceptance (Appendix~\ref{app:reproducibility}). All models and datasets used are publicly available and cited in Section~2.

\item {\bf Experimental setting/details}
    \item[] Question: Does the paper specify all the training and test details (e.g., data splits, hyperparameters, how they were chosen, type of optimizer) necessary to understand the results?
    \item[] Answer: \answerYes{}
    \item[] Justification: Section~2 presents the experimental setting in the core of the paper; full details ($K$ grids, condition counts, decoding parameters per model, parsing protocol) are given in Appendices~\ref{app:details} and~\ref{app:bpe}. No training or fine-tuning is performed, so no optimizer or split details apply.

\item {\bf Experiment statistical significance}
    \item[] Question: Does the paper report error bars suitably and correctly defined or other appropriate information about the statistical significance of the experiments?
    \item[] Answer: \answerNo{}
    \item[] Justification: Each condition aggregates 200 queries with deterministic decoding (temperature~0) and fixed sampling, so there is no run-to-run variance to report as error bars; the remaining query-sampling noise is bounded by the binomial standard error ($\le 0.05$ at $n{=}200$), six-fold smaller than the $0.3$ collapse margin. As a robustness check, the paper reports that shifting a single severity boundary by $\pm 0.05$ reclassifies at most $7\%$ of conditions and leaves the median grade unchanged (Section~2).

\item {\bf Experiments compute resources}
    \item[] Question: For each experiment, does the paper provide sufficient information on the computer resources (type of compute workers, memory, time of execution) needed to reproduce the experiments?
    \item[] Answer: \answerYes{}
    \item[] Justification: Appendix~\ref{app:details} states that all experiments ran on a single rented NVIDIA A100-40GB GPU (bf16, inference only, no training).

\item {\bf Code of ethics}
    \item[] Question: Does the research conducted in the paper conform, in every respect, with the NeurIPS Code of Ethics \url{https://neurips.cc/public/EthicsGuidelines}?
    \item[] Answer: \answerYes{}
    \item[] Justification: The research uses only publicly available models and datasets, involves no human subjects or personal data, and the submission is anonymized per the double-blind workshop policy.

\item {\bf Broader impacts}
    \item[] Question: Does the paper discuss both potential positive societal impacts and negative societal impacts of the work performed by the paper?
    \item[] Answer: \answerNA{}
    \item[] Justification: This is a foundational diagnostic study of ICL failure modes; no application, deployment, or data about people is proposed, so there is no direct societal impact to discuss. Reliable detection of query-independent predictions may indirectly improve the trustworthiness of few-shot evaluation.

\item {\bf Safeguards}
    \item[] Question: Does the paper describe safeguards that have been put in place for responsible release of data or models that have a high risk for misuse (e.g., pre-trained language models, image generators, or scraped datasets)?
    \item[] Answer: \answerNA{}
    \item[] Justification: The paper releases no models, datasets, or code at submission time and uses only publicly available assets, so no safeguard obligations arise.

\item {\bf Licenses for existing assets}
    \item[] Question: Are the creators or original owners of assets (e.g., code, data, models), used in the paper, properly credited and are the license and terms of use explicitly mentioned and properly respected?
    \item[] Answer: \answerYes{}
    \item[] Justification: All assets are cited in Section~2 with their papers and public releases; licenses, verified via the Hugging Face API, are listed in Appendix~\ref{app:details}: LLaDA-8B weights under MIT, iLLaDA-8B weights under Apache-2.0, Llama-3 (accessed via the NousResearch mirror) under the Meta Llama 3 Community License, Qwen3.5-9B-Base under Apache-2.0, XSTest under CC-BY-4.0, and SST-2 under the GLUE terms (no unified license; the original-dataset terms apply, and Socher et al., 2013 is cited).

\item {\bf New assets}
    \item[] Question: Are new assets introduced in the paper well documented and is the documentation provided alongside the assets?
    \item[] Answer: \answerNA{}
    \item[] Justification: The paper introduces no new datasets, models, or other assets; the diagnostic protocol uses existing models, and code release upon acceptance is stated in Appendix~\ref{app:reproducibility}.

\item {\bf Crowdsourcing and research with human subjects}
    \item[] Question: For crowdsourcing experiments and research with human subjects, does the paper include the full text of instructions given to participants and screenshots, if applicable, as well as details on compensation (if any)?
    \item[] Answer: \answerNA{}
    \item[] Justification: The paper involves no crowdsourcing or research with human subjects.

\item {\bf Institutional review board (IRB) approvals or equivalent for research with human subjects}
    \item[] Question: Does the paper describe potential risks incurred by study participants, whether such risks were disclosed to the subjects, and whether Institutional Review Board (IRB) approvals (or an equivalent approval/review based on the requirements of your country or institution) were obtained?
    \item[] Answer: \answerNA{}
    \item[] Justification: The paper involves no research with human subjects.

\item {\bf Declaration of LLM usage}
    \item[] Question: Does the paper describe the usage of LLMs if it is an important, original, or non-standard component of the core methods in this research? Note that if the LLM is used only for writing, editing, or formatting purposes and does \emph{not} impact the core methodology, scientific rigor, or originality of the research, declaration is not required.
    \item[] Answer: \answerNA{}
    \item[] Justification: LLMs are the object of study rather than a component of the methodology; no LLM was used to generate data, labels, or experimental results.

\end{enumerate}

\end{document}